\documentclass[11pt]{article}

\usepackage{xspace}
\usepackage{enumerate}
\usepackage{amssymb}
\usepackage{graphicx}
\usepackage{booktabs}
\usepackage{xcolor}
\usepackage{pifont}
\usepackage{pagecolor}
\usepackage{titlesec}
\usepackage{caption}
\usepackage{subcaption}
\usepackage[table]{xcolor}
\usepackage[most]{tcolorbox}
\tcbuselibrary{minted, skins, breakable}
\usepackage{minted} 
\usepackage{xcolor}
\usepackage[percent]{overpic} 
\usepackage{hyperref}
\usepackage[capitalize]{cleveref}
\usepackage[inkscapelatex=false]{svg}
\usepackage{tikz}
\usepackage[numbers]{natbib}

\usepackage[utf8]{inputenc} 
\usepackage[T1]{fontenc}    
\usepackage{url}            
\usepackage{booktabs}       
\usepackage{makecell}
\usepackage{amsfonts}       
\usepackage{nicefrac}       
\usepackage{microtype}      
\usepackage{xcolor}         
\usepackage{multirow}
\usepackage{graphicx}
\usepackage{amsmath}
\usepackage{dsfont}
\usepackage{caption}
\usepackage{subcaption}

\crefname{section}{Sec.}{Secs.}
\Crefname{section}{Section}{Sections}
\Crefname{table}{Table}{Tables}
\crefname{table}{Tab.}{Tabs.}

\usepackage[most]{tcolorbox}
\tcbuselibrary{listings, skins, breakable}
\usepackage{xcolor}
\usepackage{listings}

\definecolor{neoblue}{RGB}{179,217,242}
\definecolor{codegray}{gray}{0.15}
\definecolor{coderule}{RGB}{150,200,230}

\lstdefinestyle{neopy}{
  language=Python,
  basicstyle=\ttfamily\small,
  keywordstyle=\color{blue!70!black}\bfseries,
  stringstyle=\color{green!40!black},
  commentstyle=\color{black!60},
  numberstyle=\tiny\color{black!50},
  numbers=left,
  numbersep=8pt,
  showstringspaces=false,
  breaklines=true,
  tabsize=4,
  keepspaces=true,
}

\newtcblisting{neocodelst}[1]{
  listing only,
  listing options={style=neopy},
  colback=white,
  colframe=neoblue,
  enhanced,
  sharp corners,
  boxrule=0.8pt,
  left=8pt, right=8pt, top=6pt, bottom=6pt,
  breakable,
  title={#1},
  attach boxed title to top left = {yshift=-2mm, xshift=2mm},
  boxed title style = {colback=white, colframe=neoblue, boxrule=0.8pt},
  borderline west  = {3pt}{0pt}{neoblue!90},
}

\definecolor{tabbaseline}{rgb}{0.7, 0.85, 0.95} 
\definecolor{tabfirst}{rgb}{1, 0.7, 0.7} 
\definecolor{tabsecond}{rgb}{1, 0.85, 0.7} 
\definecolor{tabthird}{rgb}{1, 1, 0.7} 

\definecolor{rowblue}{RGB}{220,230,240}
\definecolor{myorchid}{RGB}{150,10,30}
\definecolor{myblue}{RGB}{10,30,250}
\definecolor{mygreen}{RGB}{10,120,10}

\usepackage{amsmath,amsfonts,bm}

\usepackage[margin=1in, top=1in]{geometry}
\usepackage{graphicx}
\usepackage{fancyhdr}
\usepackage{hyperref}
\usepackage{xcolor}
\usepackage{titlesec}
\usepackage{tcolorbox}
\usepackage{titling}
\tcbuselibrary{skins}

\usepackage[T1]{fontenc}
\usepackage{tgheros}
\usepackage[utf8]{inputenc}

\usepackage{amsmath,amssymb,bm}
\usepackage{newtxtext}
\usepackage{newtxmath} 

\AtBeginDocument{%
  \providecommand{\mathds}{}%
  
  \let\mathds\mathbb
}

\providecommand{\titlefont}{\sffamily\bfseries}
\providecolor{qc_darkblue}{rgb}{0.008,0.063,0.247}
\providecolor{qc_blue}{rgb}{0.164,0.164,0.914}

\titleformat{\title}
{\titlefont\LARGE\color{qc_blue}}{}{0pt}{}

\titleformat{\section}
{\titlefont\Large\bfseries\color{qc_darkblue}}{\thesection}{1em}{}

\titlespacing*{\section}{0em}{1em}{.6em}

\newtcolorbox{titlebox}{
  enhanced,
  colback=white,
  boxrule=0pt,
  opacityback=0,
  opacityframe=0,
  width=0.95\textwidth,
  center
}

\makeatletter
\newcommand{\contactinfo}[1]{\def\@contactinfo{#1}}
\makeatother
\contactinfo{}

\fancypagestyle{titlepage}{
  \fancyhf{}
  \fancyfoot[L]{\footnotesize Qualcomm AI Research is an initiative of Qualcomm Technologies, Inc.}
  \fancyfoot[R]{\footnotesize\thepage}
  
}
\renewcommand\abstract{%
    \setlength{\parskip}{.8em}
    \par
}

\title{Paper Title}
\date{February 23, 2024}
\author{Author1, Author2}
\contactinfo{\{author1, author2\}@qualcomm.com}
\newcommand{\methodname}{\textsc{{M}obile{W}an}}

\begin{document}
\thispagestyle{titlepage}
\bibliographystyle{unsrt}

\begin{tikz}[remember picture,overlay]
  \node[anchor=north east, inner sep=0pt]
    at ([xshift=5cm,yshift=5.5cm]current page.north east)
    {\includegraphics[width=10cm,angle=-75,origin=c]{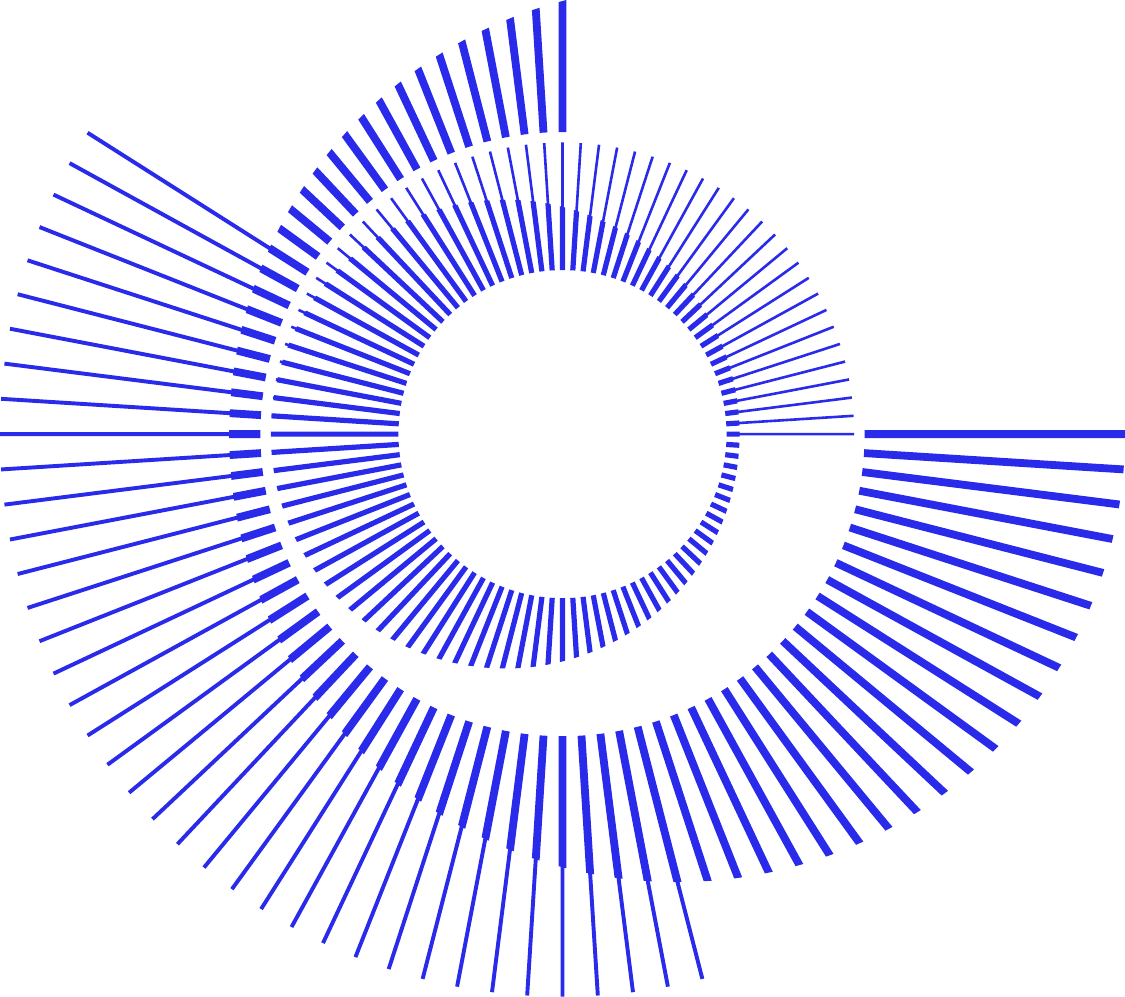}};
\end{tikz}

\begin{figure}[t]
    \vspace*{-1cm}
    \hspace*{-0.6cm} 
    \includegraphics[width=4.0cm]{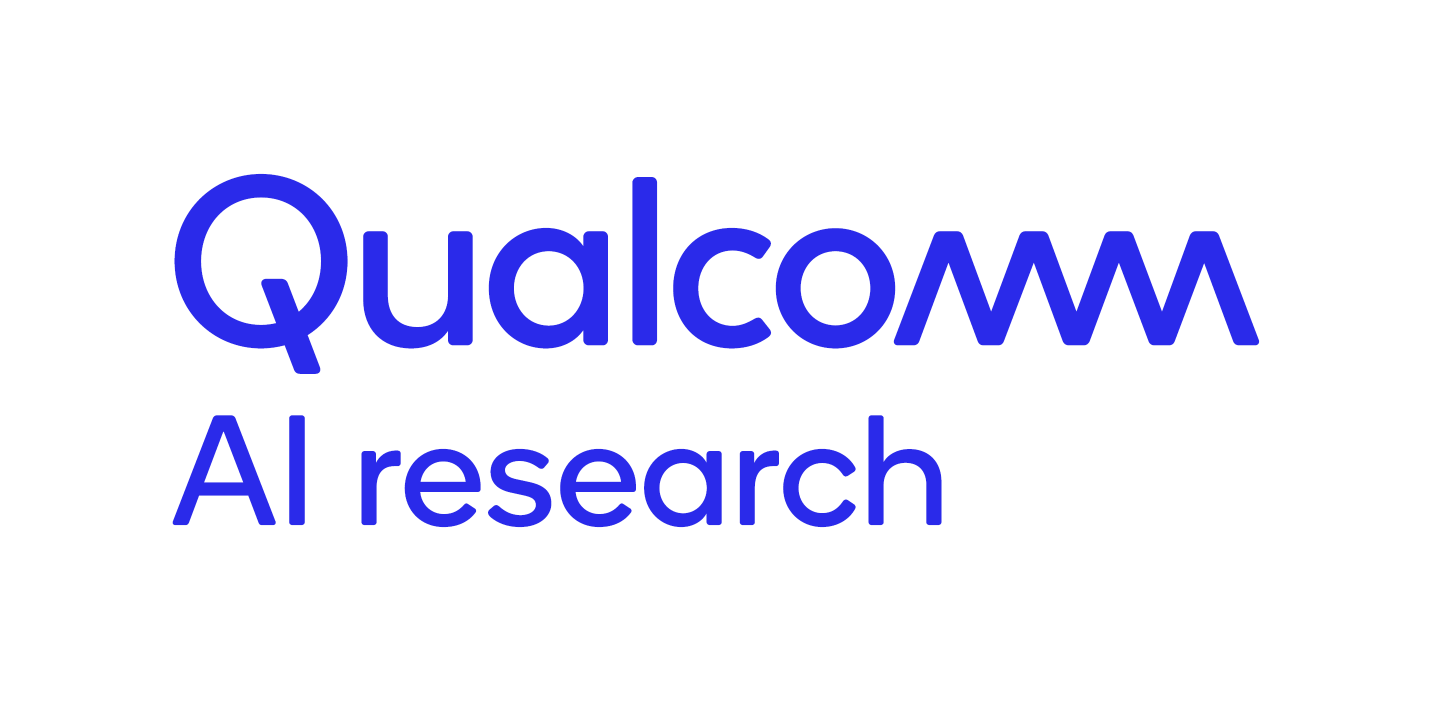}
    \vspace*{-0.5cm}
\end{figure}
\vspace{-2em}
\title{\textbf{\methodname}: \LARGE{Closing the Quality Gap for Mobile Video Diffusion} }
\date{November 6, 2025}
\author{
    Mohsen Ghafoorian\textsuperscript{*}, 
    Denis Korzhenkov\textsuperscript{*},
    Adil Karjauv\textsuperscript{*}, 
    Ioannis Lelekas\textsuperscript{*},
    Noor Fathima\textsuperscript{*},
    Spyridon Stasis, 
    Hanno Ackermann, 
    Boris van Breugel, 
    Markus Nagel,
    Fatih Porikli,
    Animesh Karnewar,    
    Amirhossein Habibian
}
\contactinfo{}

\begin{titlebox}
{\titlefont\huge\bfseries\color{qc_darkblue}\thetitle}\\

\makeatletter
{\titlefont\mdseries\normalfont\small\color{qc_blue}\@author\par
 \ifx\@contactinfo\@empty
 \else
   \vspace{-.0em}
   {\bfseries\emphfont\@contactinfo}%
 \fi}
\makeatother

\begin{center}

\centering
\includegraphics[width=\linewidth]{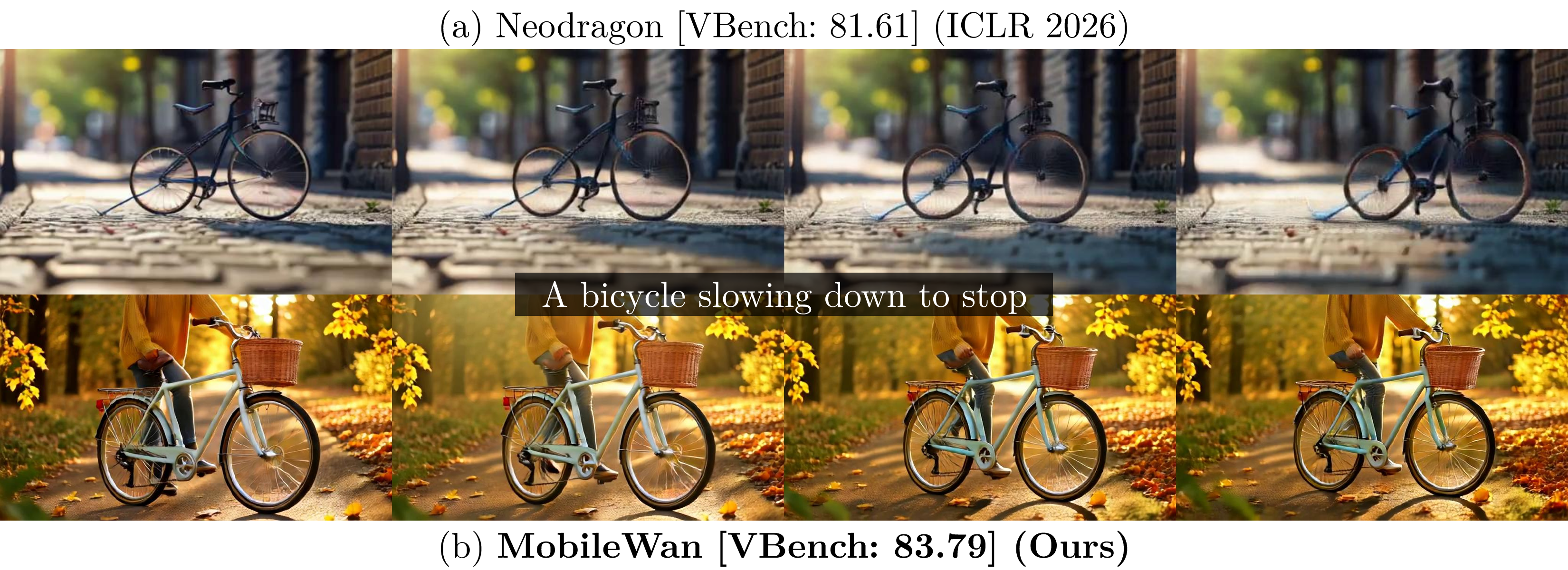}
\vspace{-1.5em}
 \captionof{figure}{We significantly advance the quality of Mobile Video Generation compared to the previous state-of-the-art (a) and produce high quality videos (b) comparable to the ones produced by server models. Our user study preferred \methodname~videos \textbf{80\%} of the time over Neodragon \cite{karnewar2026neodragon}. 
}
\label{fig:wan_vs_neodragon}
\end{center}

\vspace{-1em}

\begin{abstract}
\begin{abstract}
Recent advances in video diffusion have been driven by scaling transformer-based architectures to billions of parameters, substantially improving visual fidelity and motion coherence. In contrast, existing mobile video diffusion models remain limited to relatively small parameter budgets, typically 0.4-1.8B, restricting generation quality. In this work, we show that high-quality mobile video generation does not require small models. Instead, we demonstrate that a server-scale 5B-parameter video diffusion transformer can be deployed efficiently on memory-constrained mobile hardware through recurrent reformulation and structured compression.\\
Starting from Wan2.2-5B, we rely on a recurrence distillation framework that converts video generation into a chunk-wise autoregressive process with constant-memory attention computation. Combined with causal linear attention, the model operates as an RNN at inference time while preserving temporal coherence across chunks. We further propose a learnable attention head pruning method based on binary per-head gates optimized end-to-end using a noise-biased sparsity objective and distillation-based finetuning.
Together with sampling-step distillation and memory-optimized VAE decoding, \methodname\ becomes the first 5B-scale video diffusion model deployable on a commercial mobile device. Our system generates 5-second 480$\times$832 videos at 16 FPS in 20 seconds end-to-end latency, achieving a VBench score of 83.79 and establishing a new state of the art in mobile video generation.\\
Please find the released DiT checkpoint and the sampling code in the project page: \url{https://qualcomm-ai-research.github.io/MobileWan}
\end{abstract}

\end{abstract}


\end{titlebox}
\begingroup
\renewcommand{\thefootnote}{*}
\footnotetext{Core equal contributors}
\endgroup



\title{\methodname: \\ {C}losing the {Q}uality {G}ap for {M}obile {V}ideo {D}iffusion}

%

\section{Introduction}
Video diffusion models have rapidly become a central paradigm for generative visual modeling, enabling applications ranging from creative content production and video editing to simulation and interactive visual generation. Their recent progress has been driven by a transition toward transformer-based architectures (DiT) operating on spatio-temporal tokens, which scale effectively with data and model capacity~\cite{brooks2024videoworldsimulators,hunyuanteam2024hunyuanvideo,yang2025cogvideox,wan2025}. 

\begin{figure*}[t]
\centering
\includegraphics[width=0.98\linewidth]{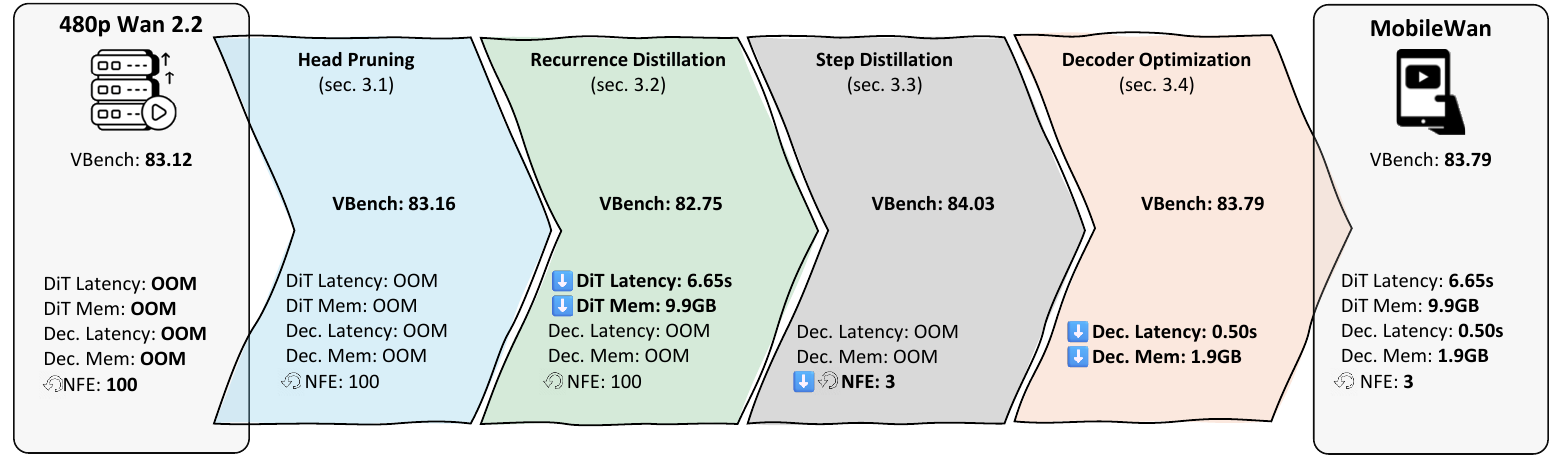}
\caption{\textbf{\methodname}: We apply a sequence of complementary optimizations, targeting both the diffusion transformer and the decoder, to the Base Wan 2.2 5B model which requires a typical Server GPU like 80GB-A100 and transform it into \methodname\ running natively on a mobile device powered by Snapdragon® 8 Gen. 5 NPU.}
\label{fig:decoder_optim}
\end{figure*}

Despite rapid progress, mobile video generation still lags substantially behind server-scale systems in terms of visual quality, temporal dynamicity/coherence and physical plausibility~\cite{shao2026efficient, bhowmik2025moalign,xue2026systematic}. Deploying video diffusion models on phones and edge devices is fundamentally constrained by peak memory consumption, the quadratic complexity of self-attention, decoder overhead, and stringent end-to-end latency requirements. As a result, existing mobile video diffusion models are typically restricted to much smaller backbones, commonly ranging from 400M to 1.8B parameters~\cite{kim2026mobile, wu2025taming, zhao2026s2dit, Ben_Yahia_2025_ICCV, wu2025snapgen, karnewar2026neodragon}. While such compact architectures satisfy the practical constraints of mobile deployment, they inevitably limit generation quality, leading to an increasingly pronounced gap between mobile and server-scale video generation.
As a concrete example, Neodragon~\cite{karnewar2026neodragon}, the current state-of-the-art video DiT for mobile, still exhibits a clear quality gap compared to server-scale models, a gap that this work seeks to bridge, as illustrated in Fig.~\ref{fig:wan_vs_neodragon}.

Bridging this gap is challenging because the bottlenecks are not confined to a single module. A natural target is self-attention, whose cost grows quadratically with the number of video tokens. However, replacing softmax attention with linear attention often either requires substantial retraining from scratch~\cite{chen2025sana} or incurs an expressiveness gap relative to full attention. More recently, works on hybrid attention~\cite{ghafoorian2026attentionsurgery, ghafoorian2026rehyat} have shown that pretrained softmax-based video diffusion models can be adapted more efficiently, narrowing the quality gap to the original model. Yet these approaches still fall short of converting the entire backbone into a fully recurrent architecture suitable for low-memory mobile deployment, leaving significant memory and compute bottlenecks unresolved.

In this work, we challenge the assumption that mobile video generation requires small diffusion models. Instead, we show that a server-scale 5B-parameter video DiT can be deployed efficiently on memory-constrained mobile hardware through recurrent reformulation and structured compression. Starting from Wan2.2-5B~\cite{wan2025}, we introduce a recurrence distillation framework that converts a pretrained transformer into a chunk-wise autoregressive model with near-constant memory attention computation. Rather than attending over the full video sequence, the model generates videos chunk-by-chunk while restricting attention to a small local token set. To preserve temporal coherence across chunks, we combine this formulation with a causal linear attention mechanism, enabling the diffusion transformer to operate as an RNN during inference.

We further introduce a learnable head pruning method tailored for video DiT's. Our approach relies on binary per-head gates optimized end-to-end with sparsity regularization and a noise-biased learning strategy that improves the pruning robustness. Combined with recurrent distillation, aggressive sampling-step distillation, and memory-optimized VAE decoding, these optimizations together enable, for the first time, deployment of a 5B-scale video diffusion model on a commercial mobile device. We present \methodname, a high-quality on-device video diffusion system built on these principles. 

Our contributions are summarized as follows:
\begin{itemize}
\item We introduce \methodname, a novel on-device video diffusion system that brings the 5B-parameter Wan2.2 model to mobile hardware, substantially narrowing the quality gap between mobile and server-scale video generation. Through system-level optimizations, we demonstrate the generation of 5-second 480$\times$832 videos at 16 fps in 20 seconds, achieving 83.79 VBench. To foster further research on mobile video diffusion, we will release our models and training recipes.

\item We show that by distilling all transformer blocks into recurrent structures, the DiT inference can be formulated as an RNN over frames, reducing the peak memory footprint as required for deploying large models on mobile devices.

\item We propose a noise-biased, learnable head-pruning method that combines per-head gating, sparsity regularization, and targeted timestep sampling, enabling more aggressive and effective pruning than heuristic alternatives under stringent compression settings.
\end{itemize}
\section{Related Work}
We review the most relevant prior work on mobile video diffusion models here, and defer the additional coverage of DiT pruning, efficient attention, step distillation, and decoder optimization to Appendix \ref{appendix_related}.

Running video diffusion models on mobile devices is severely constrained by memory, compute, and latency limitations, leading most prior work to emphasize aggressive model reductions~\citep{shao2026efficient}. Earlier on-device approaches were primarily built on UNet-based architectures~\cite{karjauv2024movie, habibian2024clockwork}. \emph{AMD-Hummingbird}~\citep{isobe2025amd} applied visual-feedback-guided pruning to reduce model size with limited quality degradation, while \emph{MobileVD}~\citep{Ben_Yahia_2025_ICCV} adapted Stable Video Diffusion~\cite{blattmann2023stablevideodiffusionscaling} through spatial downscaling, multi-scale temporal processing, and structured pruning, achieving over $500\times$ efficiency gains. \emph{SnapGen-V}~\citep{wu2025snapgen} combined temporal-layer Neural Architecture Search with adversarial step distillation for mobile deployment, and \emph{MoViE}~\citep{karjauv2024movie} targeted mobile video editing via architectural simplifications and single-step adversarial distillation. 

DiT-based VDMs have more recently begun to transition toward on-device execution. \emph{On-device Sora}~\citep{kim2025device} demonstrates training-free mobile text-to-video generation using step skipping, temporal token merging, and dynamic module loading. \emph{MVDiT}~\citep{wu2025taming} subsequently shows that mobile DiT-based video generation is feasible via extreme VAE compression, KD-guided pruning, and adversarial step distillation. Building on these directions, \emph{Neodragon}~\citep{karnewar2026neodragon} introduces a mobile-oriented pyramidal video DiT that integrates block-level pruning with extended distribution-matching and asymmetric decoder distillation, enabling 2-second video generation on mobile NPUs. Finally, \emph{S$^2$DiT}~\citep{zhao2026s2dit} presents aggressive step distillation for video diffusion transformers, serving as a general acceleration mechanism underlying many later mobile deployments. All the aforementioned mobile models are smaller than 2B parameters and exhibit notable gaps in the quality and/or size of the video they generate, with respect to the server models. Our work focuses on narrowing the \emph{quality gap} by optimizing and enabling a 5B model from Wan, a state-of-the-art and arguably the most widely used open-source video diffusion model family~\cite{wan2025}.

\section{Methods}
\subsection{Head Pruning}

Previous work on pruning diffusion transformers has focused primarily on block pruning~\citep{xie2025sana,karnewar2026neodragon,fastlightgen}. Upon an evaluation of a representative block-pruning strategy adapted from a state-of-the-art efficient video generation model~\citep{karnewar2026neodragon}, we found that Wan2.2 5B is highly sensitive to this form of compression. Please refer to the supplementary~\ref{sec:details_pruning} for more details. This observation motivates a more fine-grained pruning strategy. 

We therefore focus on pruning self-attention heads. In Wan2.2 5B, self-attention is one of the most expensive components in terms of FLOPs, which makes it a particularly relevant target for efficient, and especially on-device, generation. 
We study two approaches: a heuristic pruning method and a learned gating method.

\paragraph{Heuristic head pruning.}
Our first approach ranks heads according to an importance score and removes the least important ones. For each self-attention head, we measure its importance through its contribution to the final attention output after the output projection. More specifically, let $h_{b,i}(x_t)$ denote the output of head $i$ in block $b$, and let $W_{b,i}^{\mathrm{out}}$ be the slice of the output projection corresponding to that head. We define the head contribution \(c_{b,i}\) as
\(
c_{b,i}(x_t) = W_{b,i}^{\mathrm{out}} h_{b,i}(x_t).
\)
This can also be interpreted as the difference between the full attention output and the attention output obtained after removing head $i$. We then compute the raw importance score as the mean squared norm of this contribution:
\(
s_{b,i}
=
\mathbb{E}_{x_t,t} \|c_{b,i}(x_t)\|_2^2,
\)
where the expectation is taken over a sample set and diffusion timesteps.

Since score magnitudes may vary substantially across blocks, we normalize them within each block:
\[
\tilde{s}_{b,i}
=
\frac{s_{b,i}}{\frac{1}{H_b}\sum_{j=1}^{H_b} s_{b,j}},
\]
where $H_b$ is the number of heads in block $b$. This normalization allows scores to be compared globally across blocks. 
Given a target pruning ratio, we remove the bottom $k\%$ of heads according to this global ranking.

After pruning, we perform two-stage fine-tuning. In the first stage, we use block-wise distillation: the dense teacher and the pruned student are optimized block by block, with each block trained in isolation~\citep{chandrasegaran2026exploring}. In the second stage, we fine-tune the entire pruned model end-to-end. We refer to this procedure as the \emph{heuristic head pruning} approach.

\paragraph{Learned head pruning via gates.}
Our second approach learns which heads to prune directly from gradient signals. For each self-attention head, we introduce a learnable scalar gate. If $h_{b,i}(x_t)$ denotes the output of head $i$ in block $b$, we modulate its contribution as
\[
\hat{c}_{b,i}(x_t) = g_{b,i}\, c_{b,i}(x_t),
\qquad
g_{b,i} = \operatorname{sigmoid}\!\left(\frac{\alpha_{b,i}}{\eta}\right),
\]
where $\alpha_{b,i}$ is a learnable logit and $\eta$ is the temperature. We initialize each logit to $5.0$, so that the corresponding gate value is close to $1$, i.e., the model starts close to the dense solution. Throughout all phases of training, we jointly optimize both the gates and the model weights. This allows the remaining heads to gradually absorb useful computation from suppressed heads, leading to smoother optimization and substantially reducing the quality drop that would arise from hard pruning alone.

The model is trained with the standard flow-matching objective $\mathcal{L}_{\mathrm{FM}}$ augmented with a sparsity regularizer on the gates:
\(
\mathcal{L}_{\mathrm{pruning}}
=
\mathcal{L}_{\mathrm{FM}}
+
\lambda \sum_{b, i} g_{b,i},
\)
where $\lambda$ controls the strength of pruning. The second term explicitly encourages lower gate values, and therefore promotes sparsity. As the temperature is annealed, this regularizer drives a subset of gates to collapse toward $0$, while the remaining gates stay close to $1$, yielding an increasingly discrete pruning decision.

We optimize these gates using a three-phase fine-tuning schedule. In the first phase, we set the temperature to $\eta=1.0$, allowing the model to explore soft gate values. In the second phase, we exponentially anneal the temperature from $1.0$ to $0.1$. 

By the end of this phase, most gates are close to binary values. 
In the final phase, we binarize the gates, freeze them, and continue fine-tuning with a fixed pruned architecture under the standard noise sampling distribution.

\begin{figure*}[t]
\centering

\begin{minipage}[t]{0.49\textwidth}
    \vspace{0pt}
    \centering
    \hspace{-2em}\includegraphics[width=1.02\linewidth]{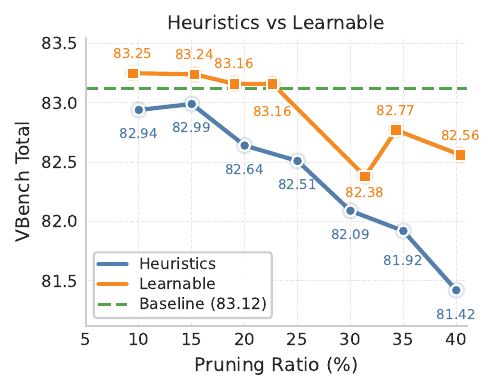}
    \vspace{-1em}
    \caption{{VBench comparison of head pruning methods.}
    Learnable pruning with high-noise-biased training performs best, especially at aggressive pruning ratios.}
    \label{fig:head_prun_heur_vs_learn}
\end{minipage}
\hfill
\begin{minipage}[t]{0.49\textwidth}
    \vspace{35pt}
    \centering
    \captionof{table}{{Effect of noise distribution during gate learning.}
    High-noise-biased training yields the best VBench score and visual quality compared with standard and low-noise-biased training.}
    \label{tab:noise_bias_head_pruning}
    \vspace{0.5em}
    \resizebox{\linewidth}{!}{
    \setlength{\tabcolsep}{3pt}
    \begin{tabular}{lccc}
        \toprule
        \multirow{2}{*}{{Model}} & \multicolumn{3}{c}{{VBench}} \\
        \cmidrule{2-4}
        & {Quality} & {Semantic} & {Total} \\
        \midrule
        Low Noise Biased   & 82.19 & 72.18 & 80.19 \\
        Standard            & 82.61 & 74.12 & 80.91 \\
        High Noise Biased  & 83.64 & 76.40 & 82.19 \\
        \bottomrule
    \end{tabular}
    }
\end{minipage}

\end{figure*}

A key observation in our experiments is that the noise distribution used during gate learning strongly affects the quality of the final pruned model. During the first two phases, we bias training toward higher noise levels by drawing $\sigma$ as $\sigma \sim \operatorname{LogitNormal}(1.5, 1)$. Intuitively, high-noise timesteps place greater emphasis on global structure and motion, and are therefore more informative for deciding which heads are essential. In the final phase, after the pruning decisions have been fixed, we return to the standard noise scheme, $\sigma \sim \operatorname{LogitNormal}(0, 1)$. We find this high-noise-biased training to be critical for successful gate learning: at a representative pruning ratio of approximately $33\%$, it outperforms training with the standard and low-noise-biased distributions in quantitative metrics (see Table~\ref{tab:noise_bias_head_pruning}).

\paragraph{Comparison and efficiency.}
Across all pruning ratios we consider, the learned gating approach consistently outperforms the heuristic approach, and the gap becomes larger at more aggressive pruning levels (see Figure~\ref{fig:head_prun_heur_vs_learn}).

Self-attention head pruning yields a measurable reduction in FLOPs. While the absolute savings are moderate compared to more aggressive structural modifications, they come with only a limited loss in quality, which makes this trade-off attractive in practice. In particular, such savings are useful in resource-constrained and on-device settings, where even modest efficiency gains can meaningfully improve deployment feasibility. Please refer to the supplementary~\ref{sec:details_pruning} for more details.


\subsection{Recurrence Distillation}
\label{subsection:rehyat}
Consider \( x \in \mathbb{R}^{N \times D} \) denoting a sequence of \( N \) tokens, and a self-attention layer \( A(x) \) represented as
\begin{equation}
A(x)=y = \text{softmax}\!\left(\frac{q k^\top}{\sqrt{D}}\right)v,
\label{eq:softmax}
\end{equation}
where queries, keys, and values are computed as linear projections as
\( q = x w_q,\; k = x w_k,\; v = x w_v \),
with learnable parameters \( w_q, w_k, w_v \in \mathbb{R}^{D \times D} \). Linear attention attempts to approximate this operation by applying the kernel trick to a rearranged formulation of the softmax attention:
\begin{equation}
y_i = \frac{\sum_{j=1}^{N} \text{sim}(q_i, k_j)\, v_j}{\sum_{j=1}^{N} \text{sim}(q_i, k_j)}
\;\approx\;
\frac{\phi(q_i)\sum_{j=1}^{N}\phi(k_j)\, v_j^\top}{\phi(q_i)\sum_{j=1}^{N}\phi(k_j)},
\end{equation}
where \( \text{sim}(q_i, k_j) = e^{q_i k_j^\top} \) denotes the unnormalized exponential similarity (up to scaling and numerical stabilization), and the approximation assumes that the exponential kernel can be factorized using a non-negative feature map \( \phi \), i.e.,
\( e^{q_i k_j^\top} \approx \phi(q_i)\phi(k_j) \).
However, since the exponential dot-product kernel has infinite rank, any feature map \( \phi(\cdot) \) that exactly reproduces the softmax attention kernel must be infinite-dimensional.

To circumvent the issue, following~\cite{ghafoorian2026rehyat}, we leverage a linear/softmax hybrid attention that benefits from the higher expressiveness of softmax and scalability of the linear attention. 
Specifically, let \( x \in \mathbb{R}^{N \times D} \) denote a flattened latent representation obtained from an underlying tensor of shape \( (T, H, W, D) \), such that \( N = T H W \).
We divide the temporal dimension into contiguous segments of \( T_c \) slices and, for each segment \( t \), define a corresponding token block
\( X_t \in \mathbb{R}^{N' \times D} \), where \( N' = T_c H W \).
Equivalently, the full latent can be viewed as a reshaped tensor
\( X \in \mathbb{R}^{T' \times N' \times D} \), with \( T' = N / N' \) denoting the number of temporal chunks; this chunk-indexed notation helps distinguish block-wise operations from single-token indexing (e.g., \( x_i \)).

Under this convention, the query matrix for chunk \( t \) is given by
\( Q_t \in \mathbb{R}^{N' \times D} \), and its corresponding linear-attention features by
\( \phi_q(Q_t) \in \mathbb{R}^{N' \times D'} \). Then for the hybrid attention output $\hat{y}_t \in \mathbb{R}^{N'\times D}$ of tokens in chunk $t$, we partition the total tokens $\mathcal{T}=\{1,2...,N\}$ to attend to, into softmax attention tokens $\mathcal{T}_t^\text{S}$ and linear attention tokens $\mathcal{T}_t^\text{L}$, and compute it as:

\begin{equation}
    \hat{y}_t = \frac{\sum_{j\in \mathcal{T}_t^\text{S}} \exp(Q_t k_j^\top / \sqrt{D} - c_t) v_j + \phi_q(Q_t) \Big( \sum_{j \in \mathcal{T}_t^\text{L}} \phi_k (k_j)\, v_j^\top \Big)}{\sum_{j \in \mathcal{T}_t^\text{S}} \exp(Q_t k_j^\top / \sqrt{D} - c_t) + \phi_q(Q_t) \Big( \sum_{j \in \mathcal{T}_t^\text{L}} \phi_k (k_j)\Big)},
\end{equation}

where \( c_t \) is a stabilizing constant (typically the maximum exponent), and \( \phi_q(\cdot) \) and \( \phi_k(\cdot) \) denote the kernel feature maps for the linear component for queries and keys.

We partition the tokens attended by chunk \( t \) into a local set
\( \mathcal{T}_t^{\mathrm{S}} = \{ j \mid tN' \le j < (t+1)N' \} \),
handled by softmax attention, and its complement
\( \mathcal{T}_t^{\mathrm{L}} = \mathcal{T} \setminus \mathcal{T}_t^{\mathrm{S}} \),
handled by linear attention.
To improve temporal coherence, we allow softmax attention to span overlapping chunks by extending
\( \mathcal{T}_t^{\mathrm{S}} \) to include \( T_o \) neighboring temporal slices, i.e. $\mathcal{T}_t^\text{S} = \{j \ | \max(tN' - T_oHW, 0) \leq j < (t+1)N'\}$, while keeping
\( \mathcal{T}_t^{\mathrm{L}} \) unchanged.

Following~\cite{ghafoorian2026attentionsurgery}, we employ learnable polynomial feature maps
\( \phi_q, \phi_k \), constructed via lightweight per-head embeddings and degree-wise expansions, to better approximate the exponential attention kernel. We first performed a block-wise distillation pretraining, where the only unfrozen parameters are the $\phi_q$s and $\phi_k$s, followed by a full DiT weights finetuning. 

\textbf{Inference-time RNN Reformulation} \\
\emph{Note that once the model is trained, at inference time, we can turn it into a full RNN thanks to the causal linear modeling of the attention to the past chunks for all the transformer blocks, enabling constant memory regardless of the sequence length}.
A convenient way to view causal linear attention is as a mechanism that maintains a small set of running aggregates. Instead of recomputing attention against the entire past at every step, one can update a fixed-size state that summarizes all previously seen keys and values.

To make such a recurrent implementation valid, the linear-attention path must be strictly causal. In our hybrid layout, causality can be enforced by choosing the temporal index split:
\begin{equation}    
    \mathcal{T}_t^\text{L} = \{j \ | j < \max(tN' - T_oHW, 0)\} 
\end{equation}
With this causal partition, sampling can be organized as a chunked recurrent procedure: the model advances in chunk steps and emits $T_c$ temporal slices per step. The softmax term acts locally within the current chunk, while all information from earlier chunks is carried forward through a compact state. Concretely, we maintain (i) a matrix accumulator $s_t \in \mathbb{R}^{D'\times D}$ that stores the running sum of feature-mapped key--value outer products, and (ii) a vector accumulator $z_t \in \mathbb{R}^{D'\times 1}$ that stores the corresponding running sum for normalization. Using these two variables, the output for chunk $t$ can be computed as
\begin{align}
    s_0 &= 0, \\
    z_0 &= 0, \\
    y_t &= \frac{\sum_{j\in \mathcal{T}_t^\text{S}} \exp(Q_t k_j^\top / \sqrt{D} - c_t) v_j + \phi_q(Q_t)\, s_t}{\sum_{j \in \mathcal{T}_t^\text{S}} \exp(Q_t k_j^\top / \sqrt{D} - c_t) + \phi_q(Q_t)\, z_t},\\
    s_{t+1} &= s_t + \sum_{j \in \mathcal{T'}_t^\text{S}} \phi_k(k_j)\, v_j^\top, \\
    z_{t+1} &= z_t + \sum_{j \in \mathcal{T'}_t^\text{S}} \phi_k(k_j),
\end{align}
where we use $\mathcal{T'}_t^\text{S}$ for updating the linear attention summaries for the next step, to account for the shifted softmax tokens boundary for the next chunk, defined as: 

\begin{equation}
    \mathcal{T'}_t^\text{S} = \{j \mid \max(tN' - T_oHW, 0) \leq j < (t+1)N' - T_oHW\}.
\end{equation}

This presentation highlights an implementation-oriented separation of concerns: intra-chunk interactions are handled by the softmax component, whereas all pre-chunk context is compressed into $(s_t, z_t)$ and accessed through the linear feature maps.

Practically, the model does not need to be optimized in this recurrent form. One may train using the equivalent causal, non-recurrent computation graph and then switch to the above stateful execution at sampling time, which yields the same outputs but enables constant-memory generation. Under this chunked-recurrent implementation, computation scales linearly with the generated sequence length, i.e., $\mathcal{O}(N)$, while the memory footprint is governed only by the fixed-size states $(s_t, z_t)$ and therefore does not grow with video duration.

 Further details on the two-stage training process is available in Appendix,~\ref{appendix_rehyat}. Table~\ref{tab:rehyat} compares the baseline Wan model with various recurrent hybrid models with different numbers of transformer blocks converted to recurrence.

\begin{table}[t]
\centering
\begin{minipage}[t]{0.49\linewidth}
\centering


\caption{\small{Comparison of VBench score and compute-burden of various hybrid recurrent models with different number of transformer blocks (out of 30) distilled to recurrence.}}
\label{tab:rehyat}
\setlength{\tabcolsep}{2.5pt}
\resizebox{1\textwidth}{!}{ \begin{tabular}{llllc}
\toprule
\multirow{3}{*}{Method} 
& \multicolumn{3}{c}{\multirow{2}{*}{VBench}} 
& \multirow{3}{*}{\makecell{Total\\Self-Attention\\TFLOPs}} \\
& \multicolumn{3}{c}{} & \\
\cmidrule(lr){2-4}
& Quality & Semantic & Total & \\
\midrule
Wan2.2 5B   & 84.16 & 78.95 & 83.12 & 43.3 \\
+ 15 Recurrent blocks & 84.79 & 79.59 & 83.75 & 37.0 \\
+ 20 Recurrent blocks & 84.51 & 80.07 & 83.62 & 34.9 \\
+ 25 Recurrent blocks & 84.07 & 80.24 & 83.30 & 32.8 \\
+ 30 Recurrent blocks & 83.68 & 80.70 & 83.09 & 26.9 \\
\bottomrule
\end{tabular} }
\end{minipage}
\hfill
\begin{minipage}[t]{0.47\linewidth}
\centering

\caption{\small{VBench scores of our optimized Wan video DiT after step distillation.}
}
\vspace{0.8em}
\label{tab:step_distillation}
\resizebox{0.8\textwidth}{!}{ \begin{tabular}{lcccc}
\toprule
\multirow{2}{*}{Method} & \multirow{2}{*}{NFEs} & \multicolumn{3}{c}{VBench}  \\
\cmidrule{3-5}
& &  Quality & Semantic & Total \\ 
\midrule
DMD   & 3 & 83.77 & 80.90 & 83.20 \\
\midrule
\multirow{3}{*}{D-DMD} &  4 & 83.41 & 80.14 & 82.75 \\%
 & 3 & 83.03 & 80.09 & 82.44 \\%
& 2 & 82.80 & 80.07 & 82.25 \\%
\midrule
\multirow{3}{*}{DMD2}  & 3 & 85.32 & 78.89 & 84.03 \\%
& 2 & 85.23 & 78.83 & 83.95 \\%
& 1 & 80.18 & 70.40 & 78.22 \\%
\bottomrule\\
\end{tabular} }
\end{minipage}
\vspace{-10pt}
\end{table}
\subsection{Step Distillation}

To further improve the latency of our system, we applied step distillation to the optimized diffusion checkpoint.
We opted for distribution matching distillation (DMD)~\cite{yin_one-step_2024} due to its proven performance for image- and video generation tasks. 
With DMD method, for the single-step student's prediction of the clean video $\hat{x}_\theta$, the forward diffusion process is applied with the noise level $\tau$, \(\hat{x}_\tau = \left(1 - \tau\right) \hat{x}_\theta + \tau \varepsilon \) with the noise \(\varepsilon \sim \mathcal{N}\left(0, I\right) \).
Given the teacher velocity model \(v\left(x_t, t\right)\) and the so-called \emph{fake-score} velocity model \(u_\varphi \left(x_t, t\right)\), gradient of the DMD loss w.r.t. the student's parameters \(\theta\) is defined as follows, \(\nabla_\theta L_{dmd} \propto \nabla_\theta L_{dmd}^{core} = \left( v^c\left(\hat{x}_\tau, \tau \right) - u_\varphi \left(\hat{x}_\tau, \tau \right) \right)^T \nabla_\theta \hat{x}_\theta \) where \(v^c\) denotes the teacher model output augmented with the classifier-free guidance (CFG)~\citep{ho_classifier-free_2021} with the scale value of $c$.
The per-sample weight of DMD loss is usually defined as $w_{dmd} = \left\Vert v^c\left(\hat{x}_\tau, \tau \right)  - \left(\varepsilon - \hat{x}_\theta \right)\right\Vert_1^{-1}$.
Fake-score model \(u_\varphi\) is trained to denoise the outputs of forward diffusion process applied to the student's generations.

Liu et al. \cite{liu2026decoupled} have recently  reconsidered the use of classifier-free guidance by the teacher model in their D-DMD work.
They proposed to split the unnormalized gradient of \(L_{dmd}^{core}\) into two terms denoted as DM (distribution matching) and CA (CFG augmentation). 
It has been shown that the main work of step distillation is conducted by the CA loss which resembles a variation of SDS loss~\citep{poole_dreamfusion_2023,mcallister_rethinking_2024} while the theoretically grounded DM loss mostly serves as a regularization term.
Namely, the teacher's output can be expressed as
\(
    v^c\left(\hat{x}_\tau, \tau \right) =   v\left(\hat{x}_\tau, \tau \right) + \left(c-1\right) \cdot \left( v\left(\hat{x}_\tau, \tau \right) - \Tilde{v}\left(\hat{x}_\tau, \tau \right) \right), 
\)
where $\Tilde{v}$ means the output conditioned on a blank or negative prompt~\citep{NEURIPS2020_49856ed4}.
Substituting this into \(\nabla_\theta L_{dmd}^{core}\) yields
\begin{align}
    \nabla_\theta L_{dmd}^{core} &= \nabla_\theta L_{DM}^{core} + \left(c-1\right) \cdot \nabla_\theta L_{CA}^{core}   
\end{align}
with \begin{align}
    \nabla_\theta L_{DM}^{core} &= \left( v\left(\hat{x}_\tau, \tau \right) - u_\varphi \left(\hat{x}_\tau, \tau \right) \right)^T \nabla_\theta \hat{x}_\theta, \\
    \nabla_\theta L_{CA}^{core} &= \left( v\left(\hat{x}_\tau, \tau \right) - \Tilde{v} \left(\hat{x}_\tau, \tau \right) \right)^T \nabla_\theta \hat{x}_\theta.
\end{align}
Instead of sharing the same noise level \(\tau\) among DM and CA terms, in D-DMD one samples different levels \(\tau_{DM}\) and \(\tau_{CA}\) as well as injected Gaussian noise terms \(\varepsilon_{DM}\)  and \( \varepsilon_{CA}\) independently.
In addition, Liu et al. \citep{liu2026decoupled} recommended using \(\tau_{CA}\) which is cleaner than the noise level \(\sigma\) of the current student's input, i.e. \(p\left(\tau_{CA} \mid \sigma \right) \propto p_{prior}\left(\tau_{CA}\right) \cdot \mathds{1}\left[\tau_{CA} < \sigma\right].\)
Sample-wise weights \(w_{term}\) for \(term \in \left\{ CA, DM \right\}\) are computed similarly to \(w_{dmd}\) but without CFG, \(w_{term} = \left\Vert v\left(\hat{x}_{\tau_{term}}, \tau_{term} \right)  - \left(\varepsilon_{term} - \hat{x}_\theta \right)\right\Vert_1^{-1}. \)
We used shifted uniform distribution~\cite{esser2024scaling} with parameter 5 as a prior for all sampled noise levels in our experiments.

In our experiments we initially conducted step distillation only with the commonly used DMD loss, following the FastVideo recipe~\cite{zhang2025faster}.
The resulting few-step generator achieves good VBench metrics with just 3 denoising steps (NFEs), see Tab.~\ref{tab:step_distillation}.
But visual analysis shows that the produced videos look slightly oversaturated to a human eye, and the dynamic degree drops in comparison with the original model (45.83 vs 68.33).
Notably, after switching to the D-DMD loss,  the color palette of the outputs changes drastically in an unfavorable way.
We found that to solve this, it is sufficient to sample $\tau_{CA}$ independently from a student's input noise level \(\sigma\) while preserving the same marginal distribution \(p\left(\tau_{CA}\right)\).
While this makes the colors look more pleasant, the amount of motion still remains lower than in case of multi-step generator. Overall, despite having lower scores, D-DMD configuration produces visually appealing samples, albeit with a slightly lower dynamic degree compared to the diffusion model.
Therefore, we used D-DMD model with 3 sampling steps for our user study.

As an alternative approach to D-DMD, we tried the DMD2 pipeline, i.e. used both  adversarial loss and \(L_{dmd}^{core}\)~\cite{yin_improved_2024}.
In our implementation of the adversarial loss we followed  FastGen~\citep{fastgen2026}: a lightweight convolutional discriminator used the features of \( \hat{x}_{\tau}\) extracted by the teacher network.
Note that in this experiment the ground-true visual data is also required while in other scenarios only text prompts are used.
As shown, DMD2 increases the dynamic degree (74.72) and results in highly competitive VBench metrics with as few as 2 sampling steps.

\subsection{Decoder optimization} \label{sec:decoder_optim}

\begin{figure}[t]
\centering

\begin{minipage}[t]{0.48\linewidth}
\vspace{0pt}
\centering

\includegraphics[
  width=\linewidth,
  keepaspectratio,
]{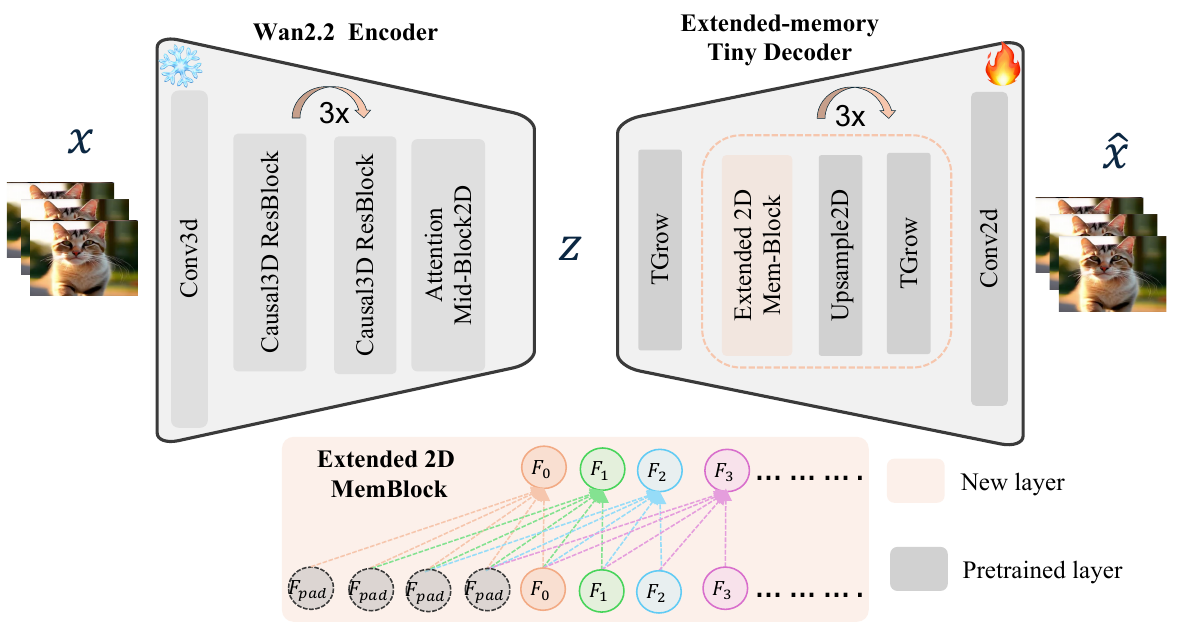}

\vspace{2pt}
\captionof{figure}{\small{Overview of the VAE architecture. We retain the Wan2.2 encoder and modify the MemBlock from LightX2V~\cite{lightx2v} by extending the causal look-back window at each layer, enabling the prediction of frame $F_t$ to leverage information from up to $F_{t-4}$.}}
\label{fig:vae_arch}
\end{minipage}
\hfill
\begin{minipage}[t]{0.48\linewidth}
\vspace{0pt}
\centering

\renewcommand{\arraystretch}{1}
\setlength{\tabcolsep}{2.5pt}

\captionof{table}{\small{Quantitative comparison of VAE variants across quality and efficiency metrics. We report VBench scores and PSNR on 60 randomly selected DAVIS~\cite{Perazzi2016,Caelles_arXiv_2018} test videos $33\times512\times512$. We also report decoder parameter counts and mobile inference latency measured at target resolution $81\times480\times832$.}}
\label{tab:vae_comparison}
\resizebox{\linewidth}{!}{%
\begin{tabular}{lcccccc}
  \toprule
  \multirow{2}{*}{{Model}} &
  \multicolumn{3}{c}{{VBench} $\uparrow$} &
  \multirow[c]{2}{*}{{PSNR $\uparrow$}} &
  \multirow[c]{2}{*}{{\shortstack{Dec.\\Params (M)}}} &
  \multirow[c]{2}{*}{{\shortstack{Mobile\\Time (ms)}}} \\
  \cmidrule(lr){2-4}
  & Total & Qual. & Sem. & & & \\
  \midrule
  Wan 2.2 VAE &  83.12 & 84.16 & 78.95 & 32.69 & 555.04 & {OOM} \\
  \midrule
  LightX2V VAE &
  82.44 & 83.54 & 78.01 &
  25.01 & 9.9 & 72.60 \\
  Tiny VAE &
  82.31 & 83.34 & 78.23 &
  26.77 & 9.9 & 72.60 \\
  {Ours} &
  {82.93} & {83.99} & {78.70} &
  {28.85} & 167.86 & 501 \\
  \bottomrule
\end{tabular}
    }
\vspace{2pt}
\end{minipage}
\vspace{-1pt}
\end{figure}

Video VAEs are commonly constructed using causal 3D convolutional operators, where the generation of a frame $F_t$ conditions on past frames (e.g., $F_{t-1}$). While effective, causal Conv3D layers introduce substantial computational and memory overhead, making them unsuitable for mobile deployment. To address this, open source approaches have developed  efficient decoder designs based on carefully structured 2D convolutions that approximate the behavior of causal Conv3D layers~\cite{lightx2v, tiny}. In this work, we build upon the open-source LightX2V \cite{lightx2v}, which is specifically trained for the Wan2.2 video model and is designed to be both compute- and memory-efficient. However, our analysis reveals that despite its efficiency, the resulting videos exhibit pronounced temporal artifacts, leading to degraded visual quality. We attribute this behavior to the limited temporal receptive field of the decoder, where each frame prediction depends on insufficient historical context.
To mitigate this issue, we modify the efficient decoder architecture by extending the causal look-back window of each layer. Specifically, the prediction of frame $F_t$ is conditioned on a longer temporal context, incorporating information from frames up to $F_{t-4}$, as illustrated in Fig.~\ref{fig:vae_arch}. This extended temporal conditioning significantly reduces temporal artifacts while preserving the efficiency advantages of the original design.

We freeze the VAE encoder of Wan2.2 and fine-tune the modified efficient decoder, as illustrated in Fig. \ref{fig:decoder_optim}. The decoder is partially initialized from the pretrained LightX2V VAE, while parameters introduced by the architectural modifications are randomly initialized. Video samples are first encoded using the frozen Wan2.2 encoder to obtain latent representations, and the decoder is trained to reconstruct the corresponding videos using a combination of an $L_2$ reconstruction loss and a perceptual loss.
We train on an in-house synthetic dataset consisting of approximately 80K videos, each video is resized to a patch size of \(480\times480\) and first 33 frames are used. The decoder is fine-tuned for 200K iterations on 8 NVIDIA A100 GPUs with an effective batch size of 8.

In Table~\ref{tab:vae_comparison}  we compare our proposed VAE against existing efficient decoder baselines and observe consistent improvements in reconstruction quality. Our architectural modifications and fine-tuning strategy lead to measurable gains across multiple metrics, including higher PSNR and improved VBench scores, compared to the efficient decoder counterparts.

\section{End-to-end Integration}
To further reduce computational cost, we first decrease the token count by downsampling the video’s spatiotemporal resolution, and then fine-tune the model at the target setting of 81$\times$480$\times$832 (16 FPS). Further details are provided in Appendix~\ref{appendix_ft}.

Table~\ref{tab:one-by-one} shows the one-by-one integration of the optimization components, i.e. head pruning, recurrent hybrid attention conversion, step distillation and decoder integration. As it can be observed, neither each of optimization components nor the whole set significantly drop the corresponding metrics. 
The detailed metrics are reported in Appendix~\ref{sec:appendix_integration}.
Table~\ref{tab:sota} compares our model to the state-of-the-art efficient video diffusion models and specifically available on-device models that report VBench~\cite{huang2023vbench}. Evidently, our model remains pretty competitive.

To enable efficient mobile deployment after model optimizations, we develop a quantization pipeline that compresses MobileWan model weights to 8-bit precision and employs mixed-precision activations, retaining sensitive activations in a few layers (e.g., residual connections) at 16 bits while quantizing the remaining activations to 8 bits. We first apply post-training quantization (PTQ) using the FastForward framework \cite{FastForward}. While PTQ significantly reduces the model's computational and memory footprint, it introduces a noticeable degradation in video quality. To mitigate this effect, we further adopt a quantization-aware training (QAT) strategy. The training is carried out in two stages. First, we perform block-wise distillation, aligning the outputs of individual quantized DiT blocks with their full-precision counterparts. We then fine-tune the entire DiT using teacher$-$student distillation, where the quantized model is trained to match the outputs of the full-precision model. This two-stage distillation process substantially improves the quality of the quantized model and reduces the accuracy gap relative to the full-precision baseline.

\begin{figure*}[t]
\centering
\includegraphics[width=\linewidth]{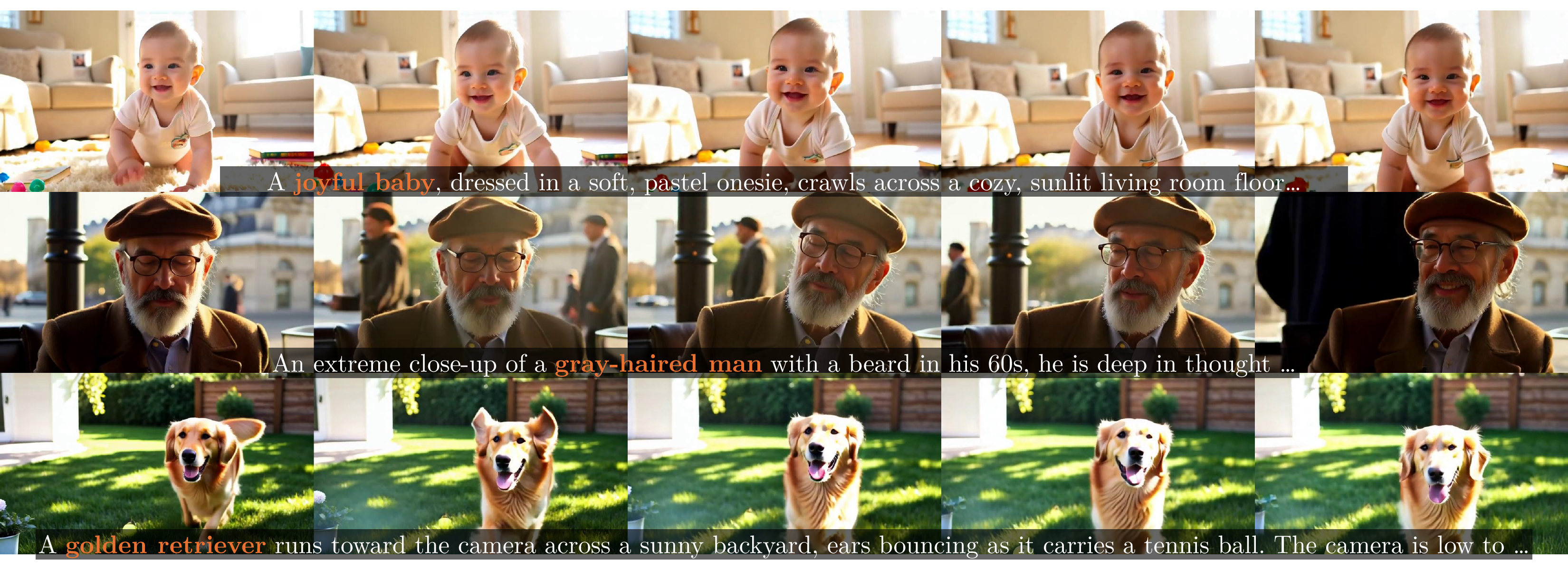}
\caption{{Qualitative Evaluation of \methodname\ model}. We visualize uniformly selected frames from the \(81\times480\times832\) videos generated on Qualcomm Snapdragon® 8 Gen. 5 NPU.
}
\label{fig:qualitative_samples}
\vspace{-10pt}
\end{figure*}

\begin{table*}[t]
\centering

\begin{minipage}[t]{0.48\textwidth}
\vspace{0pt}
\centering

\renewcommand{\arraystretch}{1.05}
\footnotesize
\setlength{\tabcolsep}{4pt}

\captionof{table}{\small{The generation quality impact of adding the optimization components one-by-one.}}
\resizebox{\linewidth}{!}{%
\begin{tabular}{llll}
\toprule
\multirow{2}{*}{Method} & \multicolumn{3}{c}{VBench} \\
\cmidrule{2-4}
                        & Quality & Semantic & Total \\
\midrule
Wan2.2 5B - Finetuned & 84.16 & 78.95 & 83.12 \\
\ \ + Head Pruning (23\%) & 84.13 & 79.30 & 83.16 \\
\ \ \ \ + Reccurence Distil. (30 Blocks) & 83.44 & 80.01 & 82.75 \\
\ \ \ \ \ \ + Step Distillation (3 steps) & 85.32 & 78.89 & 84.03 \\
\ \ \ \ \ \ \ \ + Optimized Decoder & 85.16 & 78.32 & 83.79 \\
\bottomrule
\end{tabular}
}
\label{tab:one-by-one}

\vspace{2em}

\scriptsize
\setlength{\tabcolsep}{4pt}

\captionof{table}{\small{Results of the method-blinded human visual preference study over 498 paired videos for comparison with Wan2.2 5B, and 361 for comparisons with Neodragon.}}
\resizebox{\linewidth}{!}{%
\begin{tabular}{@{\extracolsep{\fill}}lccc@{}}

\toprule
\multirow{2}{*}{Baseline} & \multicolumn{3}{c}{Human Preference \%} \\
\cmidrule{2-4}
 & \methodname & No preference & Baseline \\
\midrule
Wan2.2 5B - FT & 22 & 25 & 53 \\
Neodragon & 80 & 11 & 9 \\
\bottomrule
\end{tabular}
}
\label{tab:human_pref}

\end{minipage}
\hfill
\begin{minipage}[t]{0.48\textwidth}
\vspace{0pt}
\centering

\renewcommand{\arraystretch}{1.05}
\footnotesize
\setlength{\tabcolsep}{4pt}

\captionof{table}{\small{Comparisons with SOTA generic and on-device video diffusion models on VBench.}}
\resizebox{\linewidth}{!}{%
\begin{tabular}{lccc}
\toprule
\textbf{Up-to-5B Server Models} & Total$\uparrow$ & Quality$\uparrow$ & Semantic$\uparrow$ \\
\midrule
Open-Sora Plan V1.3~\citep{lin2024opensora} & 77.23 & 80.14 & 65.62 \\
CogVideoX 5B~\citep{yang2025cogvideox} & 81.91 & 83.05 & 77.33 \\
CogVideoX1.5 5B~\citep{yang2025cogvideox} & 82.01 & 82.72 & 79.17 \\
Open-Sora V1.2~\cite{zheng2024open} & 79.76 & 81.35 & 73.39 \\
LTX-Video~\citep{hacohen2024ltx} & 80.00 & 82.30 & 70.79 \\
Mobile Video DiT - Server~\citep{wu2025taming} & 83.09 & 84.65 & 76.86 \\
CogVideoX 2B~\citep{yang2025cogvideox} & 81.55 & 82.48 & 77.81 \\
PyramidalFlow~\citep{jin2024pyramidalflow} & 81.72 & 84.74 & 69.62 \\
M4V~\citep{huang2025m4v} & 81.91 & 83.36 & 76.10 \\
STA~\citep{zhang2025fast} & 83.00 & 85.37 & 73.52 \\
VSA~\citep{zhang2025faster} & 82.77 & 83.60 & 79.47 \\
SANA-Video~\citep{chen2025sana} & 83.71 & 84.35 & 81.35 \\
Attention Surgery~\cite{ghafoorian2026attentionsurgery} & 83.21 & 85.19 & 75.25 \\
Wan2.2 5B*~\citep{wan2025} & 83.12 & 84.16 & 78.95 \\
\midrule
\textbf{Mobile Models} & & & \\
\midrule
SnapGenV~\citep{wu2025snapgen} & 81.14 & 83.47 & 71.84 \\
Mobile Video DiT - Mobile~\citep{wu2025taming} & 81.45 & 83.12 & 74.76 \\
Neodragon~\citep{karnewar2026neodragon} & 81.61 & 83.68 & 73.36 \\
S$^2$DiT - AR~\citep{zhao2026s2dit} & 83.26 & \textbf{85.63} & 73.79 \\
\methodname (Ours) & \textbf{83.79} & 85.16 & \textbf{78.32} \\
\bottomrule
\end{tabular}
}
\label{tab:sota}

\end{minipage}
\caption{Computational efficiency of DiT reported for $1\times$ NFE generating
$81\times480\times832$ video on a Snapdragon® 8 Gen.~5 NPU.}
\label{tab:ondevice}
\scalebox{0.8}{ %
\begin{tabular}{lccccc}
\toprule
Model & TFLOPs & Params (B) & MHSA (GB) & RAM (GB) & Runtime (s) \\
\midrule
Wan2.2 5 B  & 98 & 5.0 & 15.7 & OOM & NA \\
\quad + Head Pruning  & 88 & \textbf{4.7} & 12.1 & OOM & NA \\
\quad\quad + Recurrent Distillation & \textbf{73} & 4.8 & \textbf{0.1} & \textbf{9.9} &\textbf{ 6.6} \\
\bottomrule
\end{tabular}}
\vspace{-1em}
\end{table*}

\paragraph{Computational Efficiency}
We evaluate the contribution of each optimization to the overall computational efficiency of \methodname. In addition to proxy efficiency metrics, namely the number of parameters, computational cost (TFLOPs), and attention weight size (MHSA), we report measured memory usage (RAM) and latency (Runtime) on a smartphone equipped with a Snapdragon® 8 Gen.~5 NPU.

As shown in Table~\ref{tab:ondevice} and Figure~\ref{fig:decoder_optim}, the original DiT denoiser from Wan2.2 5B fails to run on mobile hardware primarily due to its excessive memory footprint. Although head pruning yields substantial reductions in TFLOPs and attention weight size, it alone is insufficient to enable on‑device deployment. In contrast, recurrent distillation makes mobile inference feasible by restricting attention computation to a limited set of tokens, resulting in a drastic reduction in attention weights. This enables a denoising step over the entire video to be completed in 6.6 seconds.

\paragraph{Human Evaluation}
We extend our quantitative evaluation with two internal user studies. 
In the first survey, we compare our \methodname\ to the original diffusion checkpoint finetuned for our spatiotemporal resolution.
In the second, we evaluate the proposed system against the recent Neodragon pipeline~\cite{karnewar2026neodragon}.
Participants were shown a text prompt, randomly sampled from the VBench set, and two videos generated by the models compared.
They needed to choose the better video or declare that they had no preference. In total, we collected 859 paired comparisons, the results of which are summarized in Tab.~\ref{tab:human_pref}.
As shown, our optimized system in half of the cases is not performing worse than the full diffusion Wan2.2 model, while significantly outperforming the device-oriented Neodragon pipeline.
Qualitative examples are presented in Figures~\ref{fig:wan_vs_neodragon},  \ref{fig:qualitative_samples}. Further generated samples and qualitative comparison with other models are available in the appendix and supplementary.

\vspace{-8pt}
\section{Conclusion}

We presented \methodname, a high-quality on-device video diffusion model that narrows the gap to server-scale video generation under mobile constraints. Starting from Wan2.2, we combined a learning-based self-attention head pruning method with a fully recurrent linear/softmax hybrid reformulation of self-attention, together with decoder and distillation improvements, to enable efficient mobile deployment of a 5B-scale video diffusion model. Our final system generates 5 seconds of 480$\times$832 video at 16 fps with 20 seconds end-to-end latency, while achieving a VBench score of 83.79, competitive with the corresponding server-scale Wan model. Overall, our results show that large video diffusion transformers can be made practical on mobile devices through fine-grained compression and recurrent reformulation, without requiring an extreme loss in quality.
Nevertheless, our model still demonstrates some limitations. 
First, the quality of some scene types, e.g. human faces in the middle distance, is suboptimal. 
Second, some generations may demonstrate oversaturated colors or reduced motion (depending on the chosen step distillation method).
Third, in certain cases our RNN reformulation leads to temporal discontinuities in the generated scenes.
Finally, a combination of step distillation and optimized decoder causes flickering visible in some videos.
We plan to improve these aspects in future work.

\newpage
\appendix
\section{Technical appendices and supplementary material}
\subsection{Extended Related Work}
\label{appendix_related}
\textbf{DiT Pruning}.
Model pruning has emerged as one of the main tools for reducing the cost of diffusion transformers, both in image and video generation. In image generation, a number of recent works have explored structured pruning at different granularities, including block pruning, and channel- or head-level sparsification, often combined with distillation or post-pruning fine-tuning to recover quality~\citep{xie2025sana, fang2025tinyfusion, zhang2024learnable, zhu2025obs}. In the video domain, however, pruning remains relatively underexplored, and most existing approaches have focused on structured architectural reduction rather than a detailed study of fine-grained pruning behavior. In particular, recent efficient video diffusion systems, such as \citep{wu2025taming} and \citep{karnewar2026neodragon}, employ structured pruning as part of broader efficiency pipelines. \citep{wu2025taming} explores pruning at multiple levels, including blocks, attention heads, and FFN dimensions, while \citep{karnewar2026neodragon} relies on block-level pruning. While such approaches can provide substantial efficiency gains, they also make relatively rigid pruning decisions at the block level or combine several compression mechanisms at once, making it difficult to isolate the role of finer-grained pruning. Our work is motivated by the observation that whole-block removal can be too coarse for modern large-scale video diffusion transformers such as Wan2.2 5B. We therefore focus specifically on self-attention head pruning and compare both heuristic and learnable pruning strategies under the same video generation setting.

\textbf{Efficient Attention.}
Scaling video diffusion transformers to long spatiotemporal token sequences makes quadratic self-attention a major bottleneck, motivating progress along (i) structured sparsity, (ii) sub-quadratic approximations, and (iii) implementation-level acceleration.
Structured sparse patterns (e.g., local, blockwise, or global+window connectivity) reduce attention cost while maintaining or approximating long-range interactions~\cite{child2019sparse,beltagy2020longformer,zaheer2020bigbird,roy2020routing, kahatapitiya2024object}.
Complementarily, linearized, kernelized, or low-rank approximations replace full softmax attention with factorized computation to achieve sub-quadratic (often linear) scaling~\cite{wang2020linformer,kitaev2020reformer,katharopoulos2020transformers,choromanski2020performer,xiong2021nystromformer}.
In parallel, exact attention can be substantially faster in practice via I/O-aware fused kernels that reduce memory traffic and avoid materializing the full attention matrix~\cite{dao2022flashattention,dao2023flashattention2}.
Within diffusion/video generation, these directions appear as (a) linear/hybrid attention replacements~\cite{ghafoorian2026rehyat,ackermann2026hla,chen2025sana,zhang2025sla,ghafoorian2026attentionsurgery}, (b) sparse or temporal-sparse attention~\cite{shmilovich2025liteattention,peruzzo2025adaptor}, and (c) complementary token- and resolution-scheduling pipelines~\cite{liu2025astraea,jin2024pyramidalflow,ran2025tpdiff,korzhenkov2026pyramidalwan}.
In \textsc{MobileWAN}, we follow this theme by retaining a pretrained video DiT backbone while incorporating mobile-friendly efficient attention components (including ReHyAt) under tight compute budgets.

\textbf{Step Distillation}.
Turning pretrained diffusion models into few-step generators, often referred to as `step distillation' is crucial for creating efficient generative models.
While most of the methods for such optimization are still being developed and probed on image models~\citep{10.1145/3680528.3687625,sabour2026align,Chen_2025_ICCV,cheng2026twinflow}, recent works have demonstrated their applicability to video generators as well.
SF-V~\citep{zhang2024sfv} and MobileVD~\citep{Ben_Yahia_2025_ICCV} pipelines sucessfully applied latent adversarial distillation to an image-to-video model, while APT~\citep{pmlr-v267-lin25m} scaled it to text-to-video generators.
Arguably, most of the current works rely on the DMD~\citep{yin_one-step_2024,luo_diff-instruct_2023} method that aims to match the distribution of multi-step teacher's outputs with the distribution of data points produced by a single-step student model~\citep{fastlightgen,fan2026phaseddmdfewstepdistribution,shao2025magicdistillationweaktostrongvideodistillation,nie2026transitionmatchingdistillationfast,qiu2025histreamefficienthighresolutionvideo}.
Note that in practice a few-step sampler is typically used, since getting acceptable quality with a single step at inference time is still challenging.
It is noteworthy that DMD serves as an additional loss in other methods such as rCM~\citep{zheng2026large} which adapted continuous consistency distillation to video generation.
In our work we also opted for DMD pipeline to reduce the number of denoising steps in the final system.

\textbf{Efficient Decoders}.
Latent video diffusion models employ video VAEs to compress frames into low-dimensional latent codes for video generation. Early approaches simply extended 2D image VAEs (e.g. Stable Diffusion) with causal Conv3D layers in the decoder, so each frame is generated using only past context. While these decoders preserve causal ordering and reduce flicker, they introduce substantial compute and memory overhead and do not fully exploit temporal redundancy. Open-source latent diffusion models like Open-Sora~\cite[v1.2]{zheng2024open} and NVIDIA’s Cosmos \cite{cosmos} factorize video latents into separate spatial and temporal modules, effectively replacing heavy 3D convs with chains of 2D convs for improved efficiency.  LeanVAE replaces learned upsampling with wavelet-based multiscale decoding\cite{leanvae}, and WF-VAE \cite{wfvae} integrates Haar wavelet decomposition to offload low-frequency content from the neural decoder. These designs reduce FLOPs and memory footprints (often by one or two orders of magnitude) while maintaining reconstruction fidelity.
Another challenge is maintaining temporal coherence under high compression. Recent designs extend the decoder’s temporal receptive field to mitigate artifacts. IV-VAE \cite{ivvae} uses group causal convolution, allowing a look-back window at each layer. This yields smoother frame-to-frame quality without sacrificing the ability to process long videos sequentially. Similarly, Wan2.2 VAE \cite{wan2025}  employs overlapping frame blocks or caching strategies to preserve continuity across chunked decoding. Hardware-aware decoders also emerged. LTX-Video \cite{hacohen2024ltx} and Turbo-VAED \cite{turbovae} specifically target edge devices: they shrink decoder networks via low-rank or depthwise convs and eliminate mobile-unfriendly operations like 3D pixel-shuffle, often by knowledge distillation from a larger model. Lastly, CV-VAE \cite{cvvae} focuses on latent-space compatibility i.e., it adds a regularization term to align the video VAE’s latent distribution with an existing image VAE’s space, enabling direct fine-tuning from pretrained image diffusion models. This approach ensures efficiency in terms of training cost (reusing weights and data), though not directly addressing runtime speed.

\subsection{Details on Model Pruning}
\label{sec:details_pruning}
\textbf{Block pruning vs head pruning.} 
When we applied a block-pruning strategy adapted from \emph{Neodragon}~\citep{karnewar2026neodragon} to Wan2.2 5B, we found that the model is highly intolerant to this type of compression. Following~\cite{karnewar2026neodragon}, we first estimate the importance of each transformer block. However, instead of using the original scoring procedure, we define block importance based on the discrepancy between the full model prediction and the prediction obtained when a single block is skipped. Specifically, for each block, we measure the mean squared error between the full-pass velocity prediction (with no block skipped) and the skip-pass velocity prediction (with only that block skipped), averaged over diffusion timesteps and a calibration set. Intuitively, a larger error indicates that removing the block causes a greater deviation from the original model behavior, and therefore suggests higher block importance.

We then rank blocks according to this importance score. When selecting blocks to prune, we additionally enforce a non-neighboring constraint, i.e., pruned blocks are not allowed to be adjacent, in order to reduce the risk of excessive damage to the model. After removing the selected blocks, we fine-tune the pruned model in two stages. In the first stage, we perform end-to-end fine-tuning to recover the quality lost due to block removal. In the second stage, we continue fine-tuning with additional teacher supervision, including both output distillation and intermediate representation distillation, following~\cite{karnewar2026neodragon}.

The resulting VBench scores are reported in Table~\ref{tab:block_pruning_vbench_table}. As the results show, quality degrades rapidly as more blocks are removed. Pruning three blocks still gives marginally acceptable quality, but at a comparable FLOPs reduction, self-attention head pruning yields noticeably better performance. Based on this observation, we focus on self-attention head pruning in the main paper.

\begin{table}[t]
\centering

\begin{minipage}[t]{0.49\linewidth}
    \vspace{0pt}
    \centering
    \includegraphics[width=0.9\linewidth]{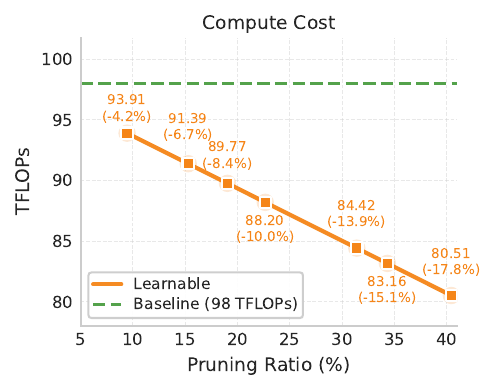}
    \captionof{figure}{{Head Pruning FLOPs vs Pruning Ratio.}
    Number of FLOPs and the corresponding reduction at different pruning ratios with respect to the baseline.}
    \label{fig:head_prun_flops}
\end{minipage}
\hfill
\begin{minipage}[t]{0.49\linewidth}
    \vspace{5pt}
    \centering
    \caption{{Block Pruning VBench Scores.}
    VBench scores for block-pruned models after Stage 1 and Stage 2 (following~\cite{karnewar2026neodragon}) with different numbers of removed blocks.}
    \label{tab:block_pruning_vbench_table}
    \resizebox{\linewidth}{!}{%
        \begin{tabular}{ccccccccccc}
\toprule
\multirow{2}{*}{{Stage}} & \multirow{2}{*}{{\# Blocks Pruned}} & \multicolumn{3}{c}{{VBench~$\uparrow$}} \\
\cmidrule{3-5}
 & & {Quality} & {Semantic} & {Total} \\
\midrule

\multirow{4}{*}{1} & 10  & 81.24 & 69.28 & 78.85 \\
& 8 & 82.66 & 76.07 & 81.34 \\
& 5 & 82.87 & 78.25 & 81.95 \\
& 3 & 83.51 & 78.27 & 82.46 \\

\midrule

\multirow{4}{*}{2} & 10 & 81.25 & 68.83 & 78.76 \\
& 8 & 82.04 & 74.73 & 80.58 \\
& 5 & 82.62 & 76.66 & 81.43 \\
& 3 & 83.12 & 77.29 & 81.95 \\

\bottomrule
\end{tabular}
    }
\end{minipage}

\end{table}

\textbf{Self-attention head pruning FLOPs gain.} Figure~\ref{fig:head_prun_flops} shows the FLOPs reduction at different head pruning ratios. The computational cost decreases steadily as more self-attention heads are removed. Although the savings are moderate, they are achieved with relatively limited quality loss, making head pruning attractive for resource-constrained and on-device deployment.
\subsection{Details on recurrent hybrid attention}
\label{appendix_rehyat}

\subsubsection{Two-stage Training}
We follow the efficient training recipe from~\cite{ghafoorian2026attentionsurgery}, and break the training into two-stages of block-wise attention distillation and light-weight finetuning.

\textbf{Block-wise recurrence distillation}. The goal here is to independently distill the bidirectional softmax attention of each transformer block into the corresponding student recurrent hybrid attention. For a smpled  prompt $p$, noise $\epsilon$ and denoising iteration $i$, the objective is to minimize the $L_1$ difference between the student and teacher attention outputs.
\begin{equation}
\boldsymbol{\phi}^l = \boldsymbol{\phi}^l - \eta\, \nabla_{\boldsymbol{\phi}_l} \Bigg(
\mathbb{E}_{\substack{\epsilon \in \mathcal{N} \\ p \in \mathcal{P} \\ i \in \mathcal{S}}}
\big|y^{(l, \epsilon, p, i)} - \hat{y}^{(l, \epsilon, p, i)}\big|
\Bigg).
\end{equation}
As specified above, in this phase we train only the linear attention kernels of each layer $l\in\{0,...N\}$ given by $\phi^l=(\phi_q^l, \phi_k^l)$. 
More specifically, we pre-train the linear attention kernels for 30k iterations with the AdamW optimizer with a learning rate of 1e-3. 

\textbf{Block selection}. We observed that converting the first few blocks have a more significant impact on the quality of the model. Therefore, we simply avoid converting the first 3 blocks in variations with lower than 27 blocks, i.e. 15, 20, and 25 blocks. More specifically, for the variant targeting $n$ recurrent blocks, we hybridize the blocks in the range $ [\ \text{min}(3, 30-n), \text{min}(3, 30-n) + n)$.

\textbf{Hybrid Model Architectural Details}. The final model we select for the on-device porting and evaluation has the following architectural specifications: All 30/30 blocks are converted to recurrence, resulting in a full RNN graph. Two-layer MLPs with polynomial degree of 3 is used for the $\phi_q$ and $\phi_k$ linear attention kernels. Recurrence latent chunk size $T_s$ is 4 and chunk overlap $T_o$ is set to 2. 

\begin{figure*}[!t]
\centering
\includegraphics[width=0.7\textwidth]{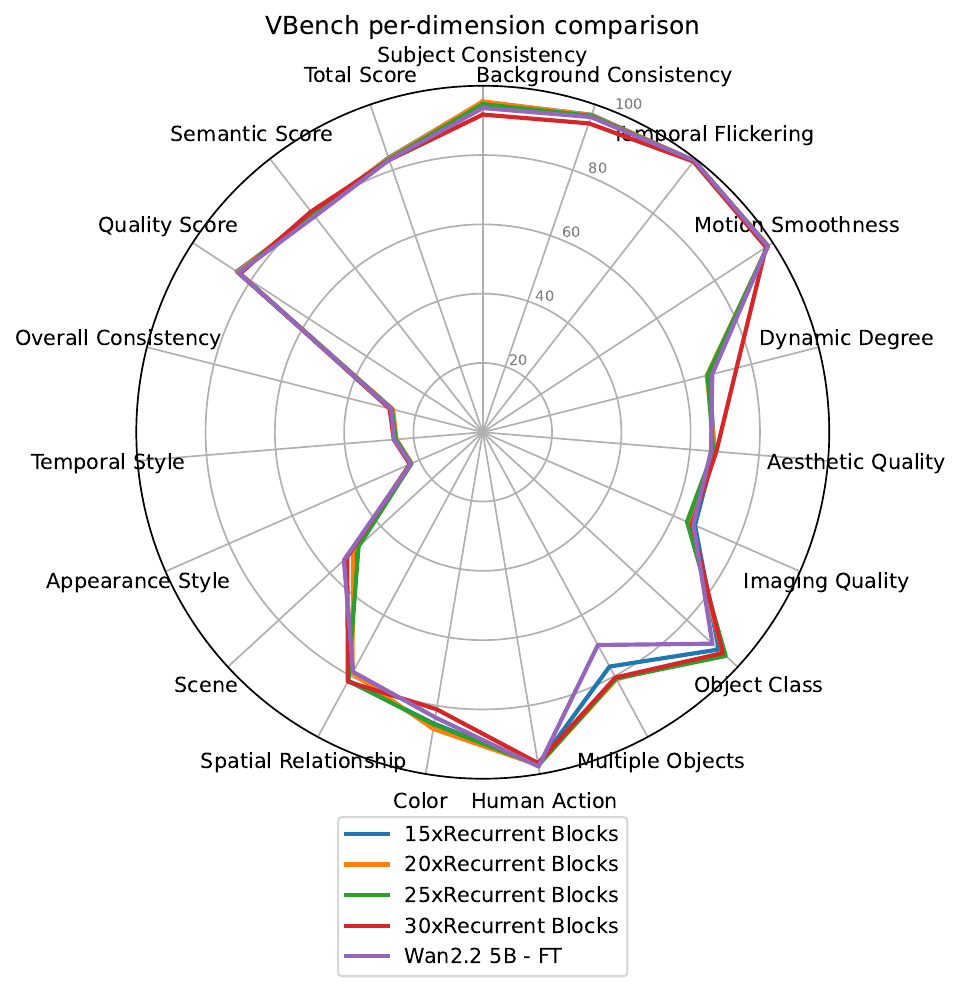}

\caption{Radar plot comparing a subset of our hybrid models with the original Wan2.2 5B model on the full VBench set and 480$\times$832 resolution}
\label{fig:rehyat_radar}
\end{figure*}

\textbf{Light-weight Finetuning}. After pretraining the linear attention kernels, we finetune the whole DiT model for 10k iterations with the flow-matching loss with the AdamW optimizer with a mini-batch size of 16 and learning rate of 4e-5, on a dataset of 80k videos synthetically generated by Wan2.1 14B. Figure~\ref{fig:rehyat_radar} provides more details on the per-dimension VBench comparison of the various recurrent hybrid model configurations to the finetuned Wan2.2 5B model. Note that the notably higher dynamic degree metric for full-RNN model (30 converted blocks) appears to be due to higher temporal jitters. The following step for teacher-based DMD step-distillation and finetuning resolves the artifacts observed.

\subsubsection{Overall Architecture}
\begin{figure*}[t]
\centering
\includegraphics[width=\linewidth]{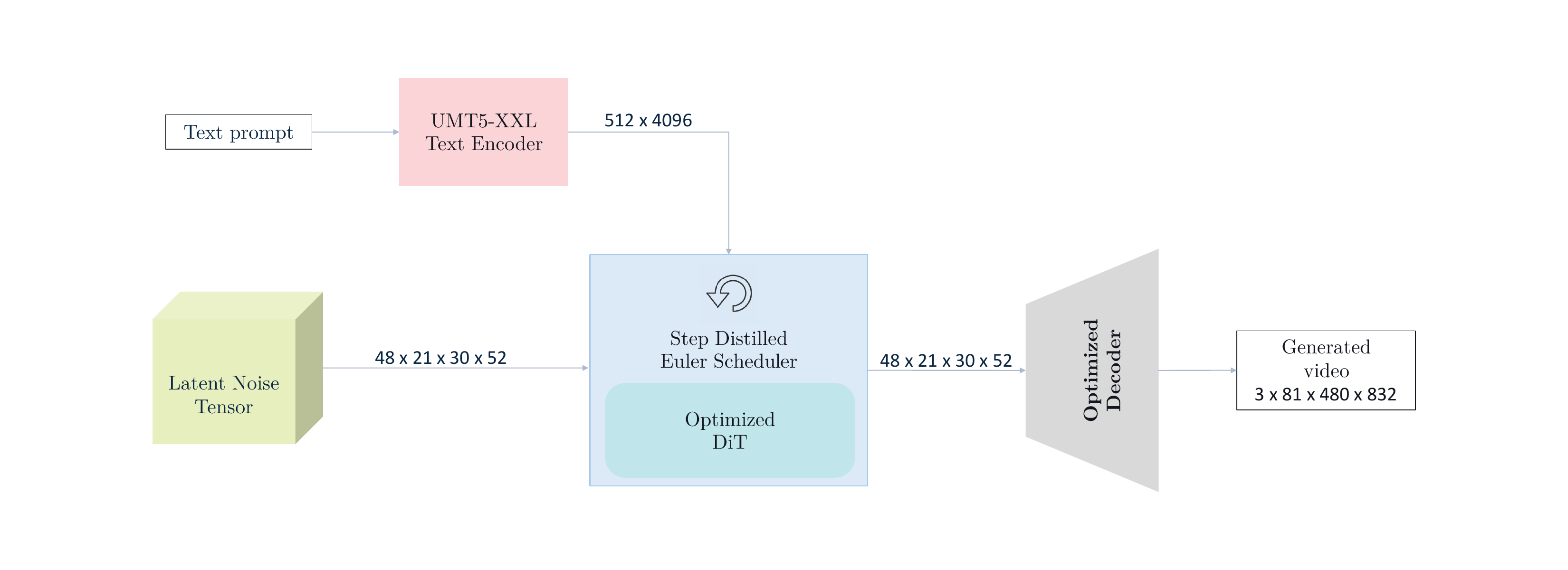}
\caption{
    Our optimized pipeline matches exactly the base Wan 2.2 model's instantiation of the latent video diffusion framework, but replaces the DiT and the Decoder components with our optimized versions. 
}
\label{fig:e2e_pipeline}
\vspace{-10pt}
\end{figure*}

Figure \ref{fig:e2e_pipeline} describes our final optimized pipeline which still remains an instantiation of the latent video diffusion framework at the core, but replaces the Scheduler, DiT and the Decoder components with our optimized versions.

\subsection{Target Resolution Finetuning}
\label{appendix_ft}
To reduce the number of tokens for our DiT, we opt to bring down the spatio-temporal resolution of the videos from 121$\times$704$\times$1280 (24 FPS) to 81$\times$480$\times$832 (16 FPS). However, the original Wan2.2 5B model, unlike the 14B model, does not support the 480p resolution, resulting in noticeably subpar generation quality and shaky camera artifacts, perhaps partially due to mismatched RoPE embeddings. Moreover, sampling 81 frames instead of 121 on the released checkpoint, results in lower dynamic contents as it resembles stretching a 3.3 second generation into 5 second by just accordingly decreasing the frame-rate. To alleviate these issues, prior to applying our optimization steps, we finetuned the model on a dataset of 5-second 480p videos at 16 FPS. The dataset consists of 80K videos synthetically generated by Wan2.1 14B. The model was finetuned for 2k iterations with the AdamW optimizer with a batch-size of 6 and learning rate of 1e-5, minimizing the flow-matching loss with a logit normal timestep distribution. Table ~\ref{tab:finetuning} shows the impact of the finetuning step in terms of quality and compute.

\begin{table}[!h]
\centering
\renewcommand{\arraystretch}{1.05} 
\footnotesize
\caption{Comparison of generation quality of the original Wan2.2 5B model before and after finetuning for our target resolution.}
\label{tab:finetuning}
\begin{tabular}{@{}lccccc@{}}
\toprule
\multirow{2}{*}{Method} & \multirow{2}{*}{Resolution} & \multirow{2}{*}{\#Tokens} & \multicolumn{3}{c}{VBench} \\ \cmidrule{4-6}
                        & & & Quality & Semantic & Total \\ \midrule 
Original Wan2.2 5B  & 121$\times$704$\times$1280 &  27280   &   85.03  &  76.28  & 83.28        \\
Original Wan2.2 5B         & 81$\times$480$\times$832  &  8190   & 82.46   &  76.64      & 81.30   \\
Wan2.2 5B - Finetuned         & 81$\times$480$\times$832  &  8190    & 84.16   & 78.95    & 83.12\\

\bottomrule \\
\end{tabular}
\end{table}
\subsection{Qualitative Results}
\begin{figure}[h]
\centering
\includegraphics[width=0.9\linewidth]{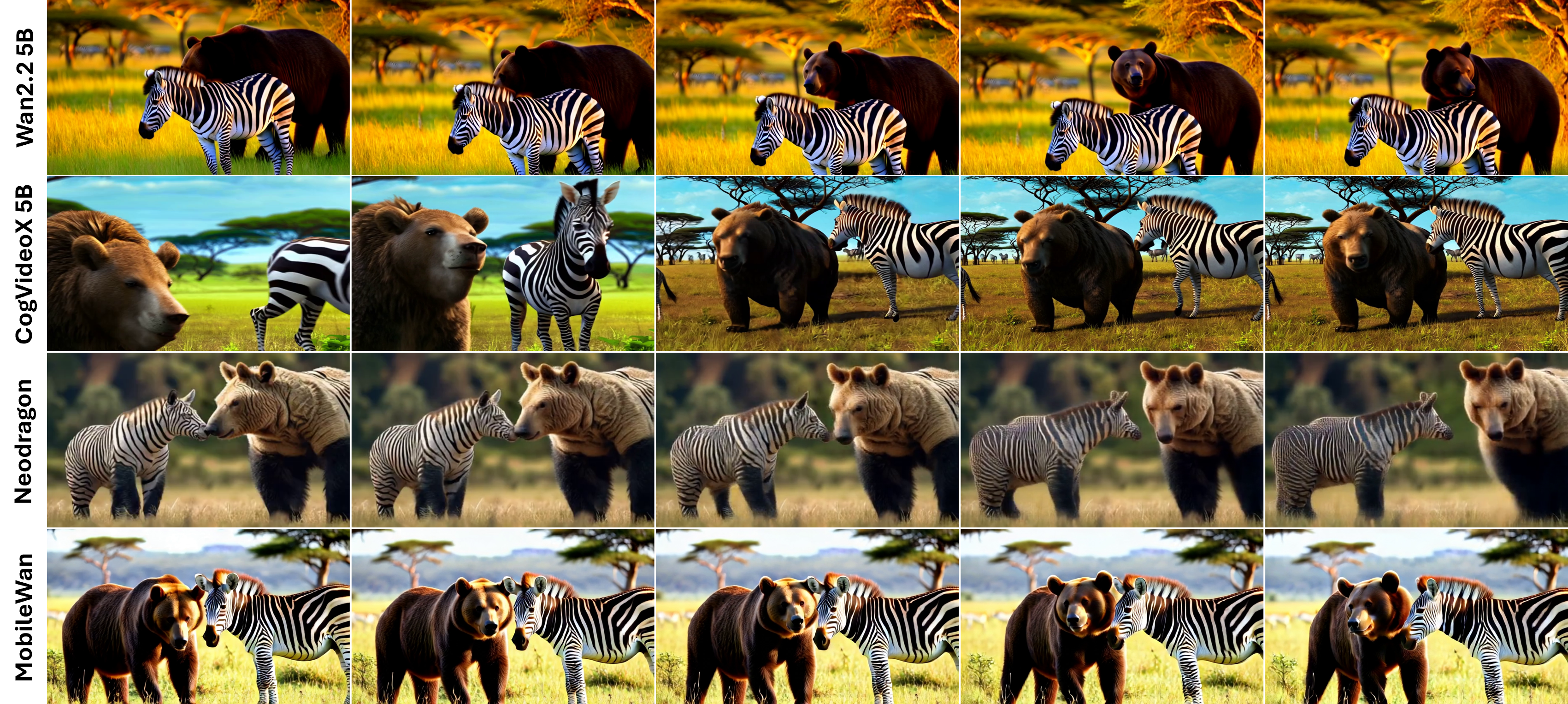}
\caption{\textbf{Prompt:} A bear and a zebra.}
\label{fig:qual_res_supp_00}
\vspace{-0.5em}
\end{figure}

\begin{figure}[h]
\centering
\includegraphics[width=0.9\linewidth]{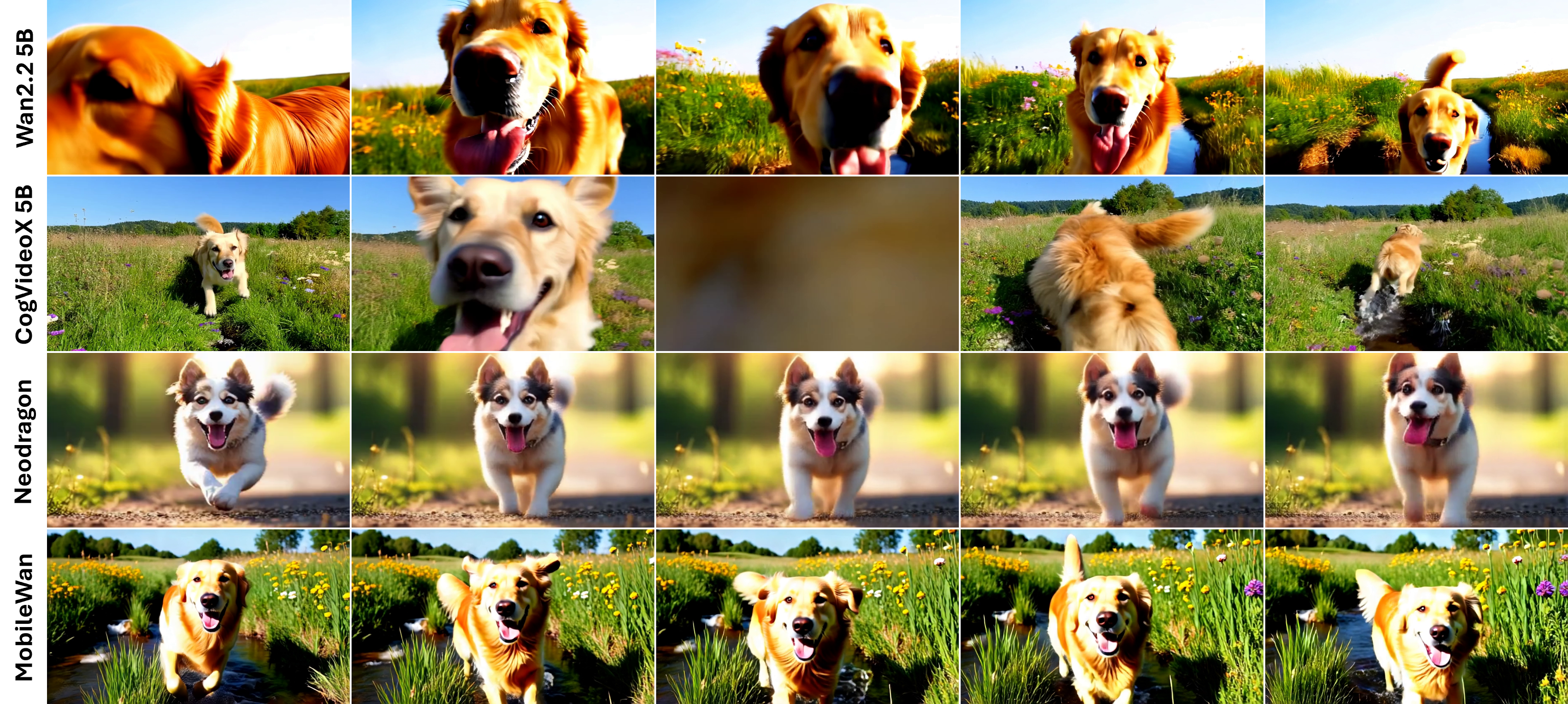}
\caption{\textbf{Prompt:} A dog running happily.}
\label{fig:qual_res_supp_01}
\vspace{-0.5em}
\end{figure}

\begin{figure}[h]
\centering
\includegraphics[width=0.9\linewidth]{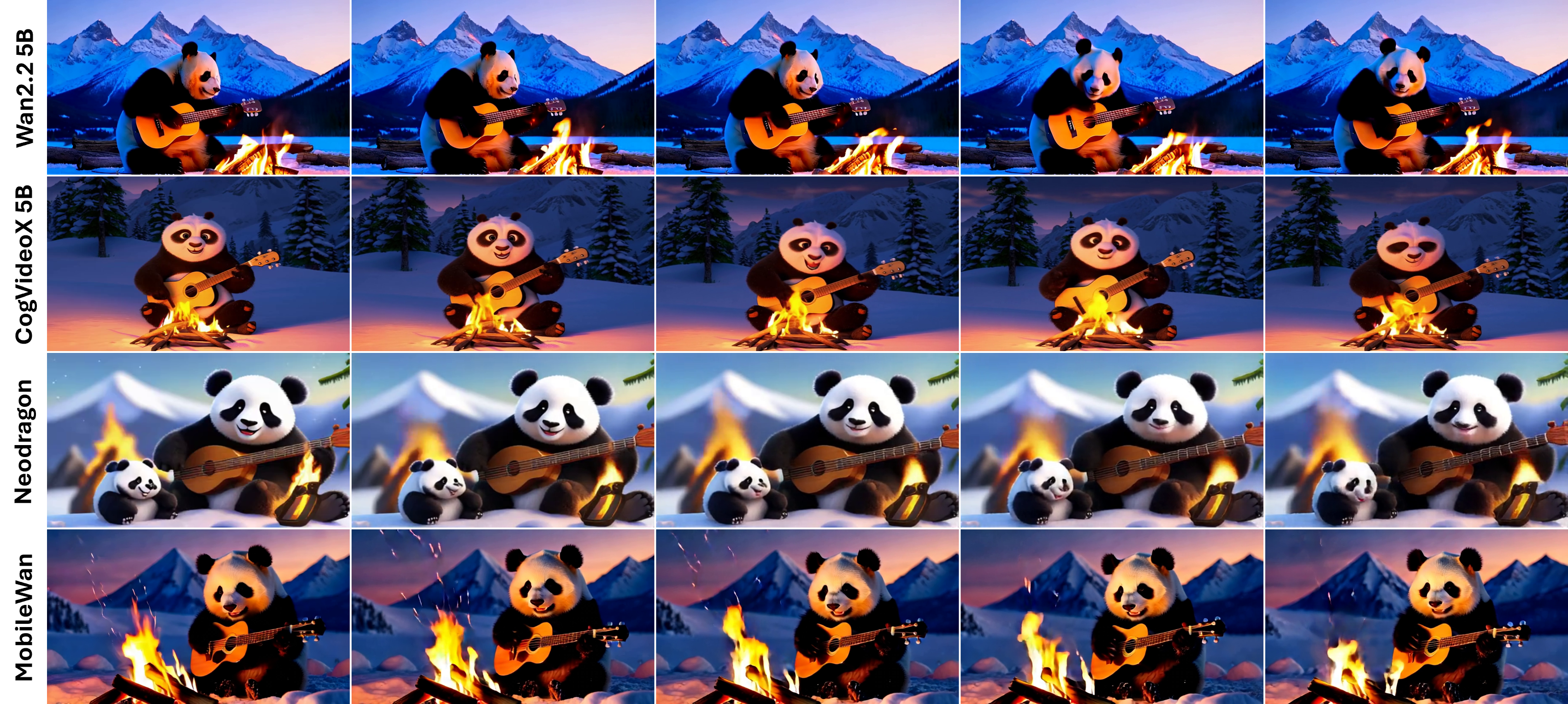}
\caption{\textbf{Prompt:} \small{A happy fuzzy panda playing guitar nearby a campfire, snow mountain in the background.}}
\label{fig:qual_res_supp_02}
\vspace{-0.5em}
\end{figure}

\begin{figure}[h]
\centering
\includegraphics[width=0.9\linewidth]{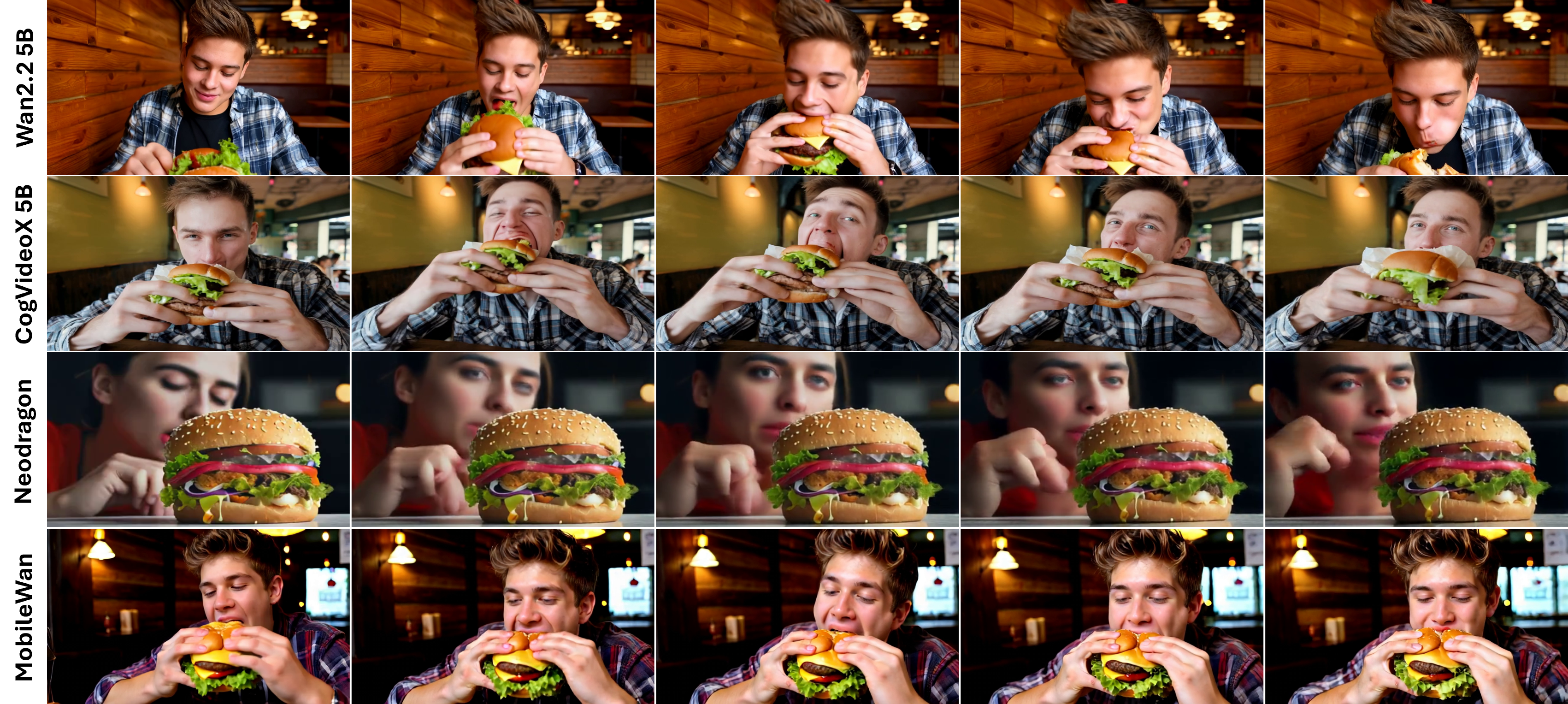}
\caption{\textbf{Prompt:} A person eating a burger.}
\label{fig:qual_res_supp_03}
\vspace{-7.5em}
\end{figure}

\begin{figure}[h]
\centering
\includegraphics[width=0.9\linewidth]{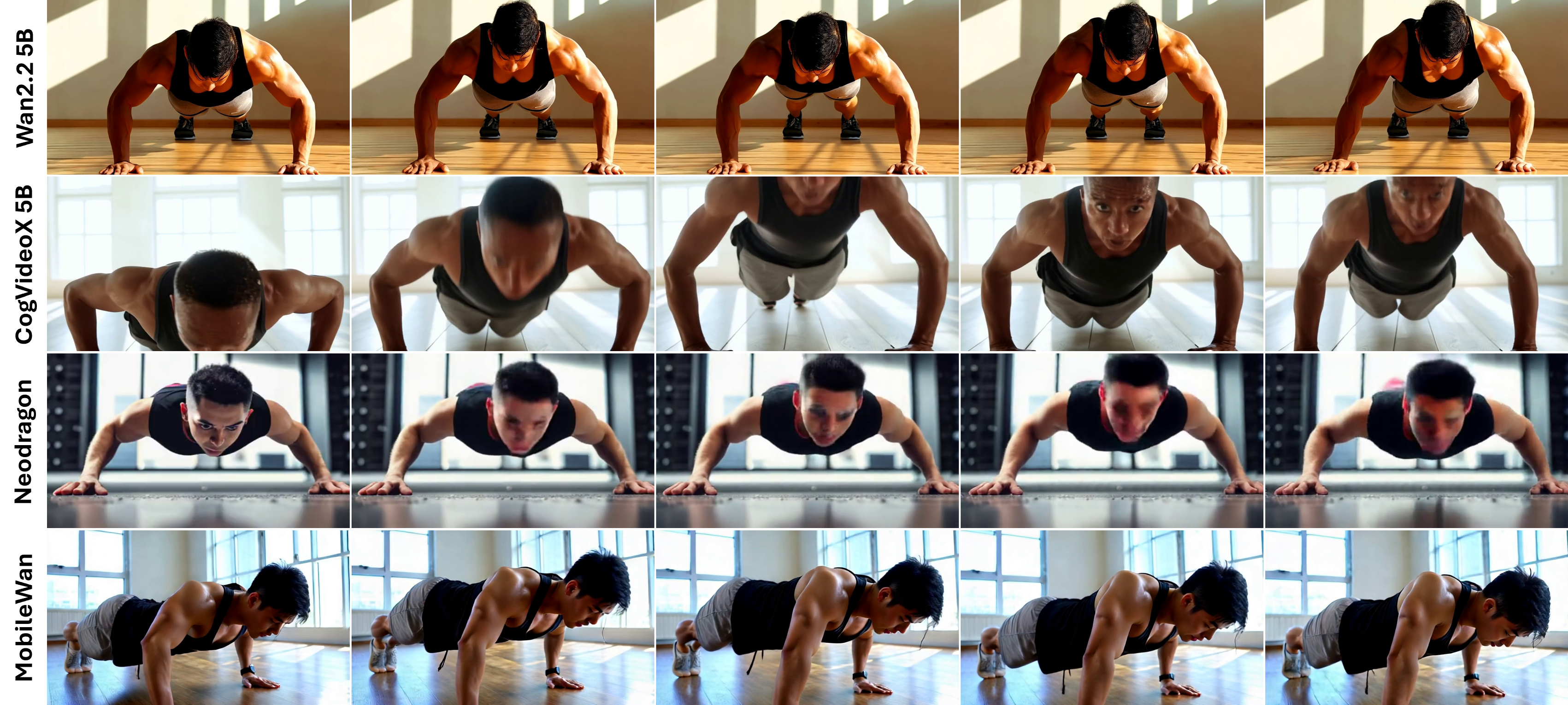}
\caption{\textbf{Prompt:} A person is pushing up.}
\label{fig:qual_res_supp_04}
\vspace{-7.5em}
\end{figure}

\begin{figure}[h]
\centering
\includegraphics[width=0.9\linewidth]{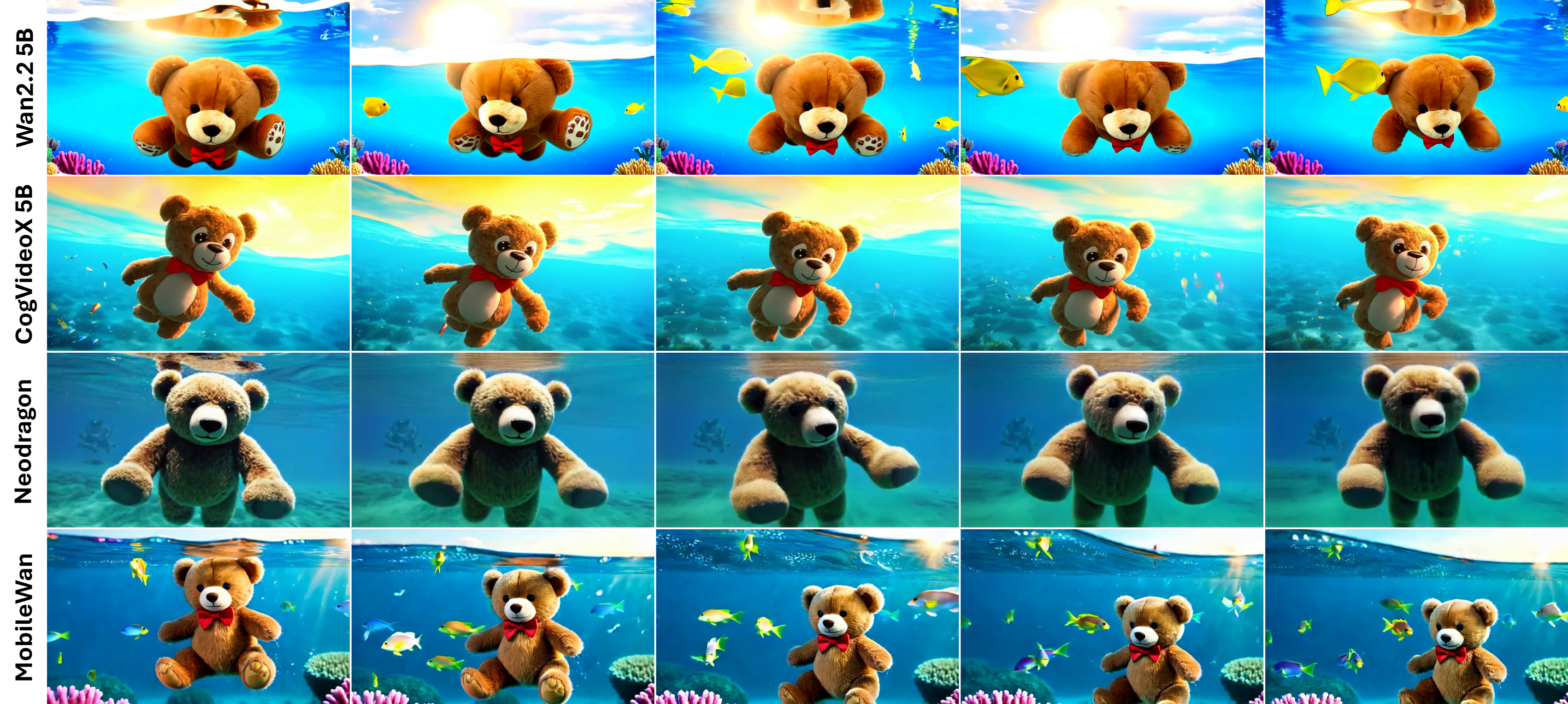}
\caption{\textbf{Prompt:} A teddy bear is swimming in the ocean.}
\label{fig:qual_res_supp_05}
\vspace{0em}
\end{figure}

We include additional qualitative comparisons here (see Figures~\ref{fig:qual_res_supp_00},\ref{fig:qual_res_supp_01},\ref{fig:qual_res_supp_02},\ref{fig:qual_res_supp_03},\ref{fig:qual_res_supp_04}, \ref{fig:qual_res_supp_05}). Note that \methodname\ generates videos of competitive or better quality than previous state-of-the-art server and mobile models of comparable size. Please refer to the supplementary for the videos.

\subsection{Details on end-to-end integration}
\label{sec:appendix_integration}

\begin{figure*}[t]
\centering
\includegraphics[width=0.9\linewidth]{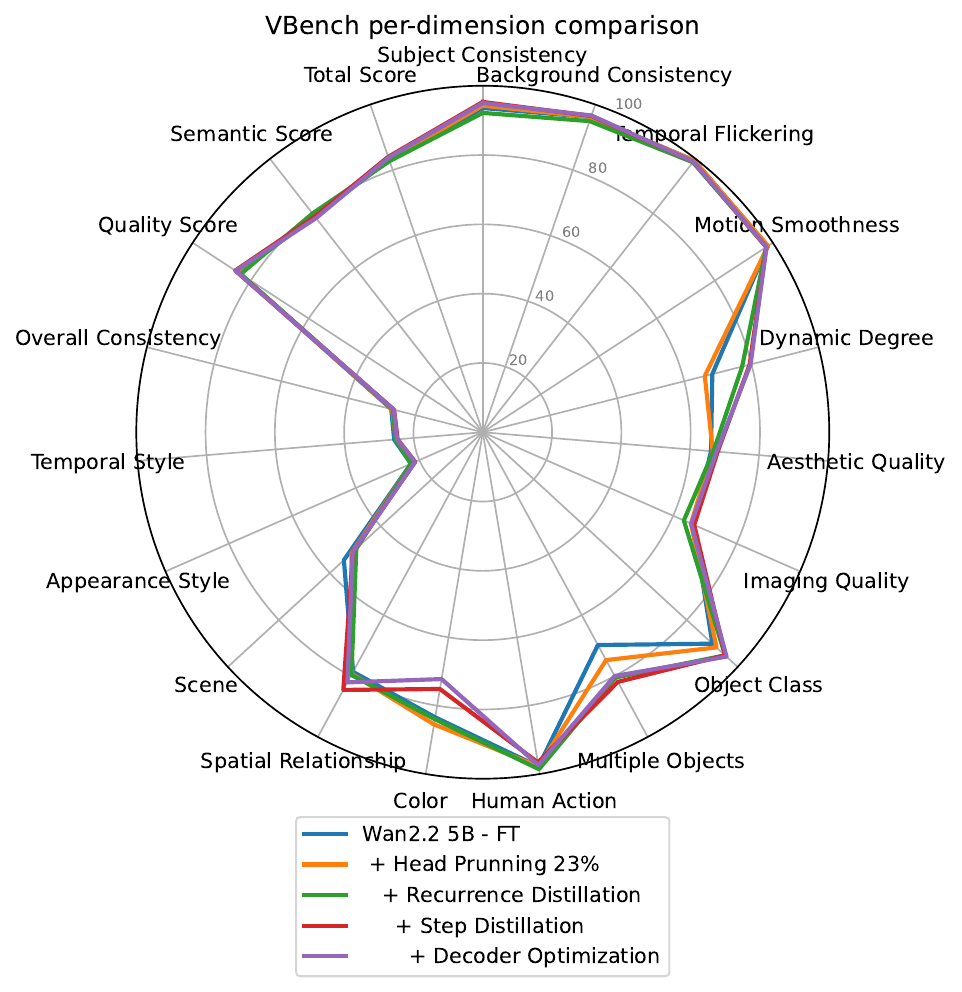}
\caption{Full set of VBench dimensions for optimizations steps used by \methodname.
}
\label{fig:integration_radar}
\end{figure*}

We provide an extended version of Table~\ref{tab:one-by-one} in Figure~\ref{fig:integration_radar} to illustrate the impact of every optimization step.
While step distillation introduces some drop in the `Color' dimension, this does not affect the visual appeal of the produced videos.
In fact, we found in our experiments that VBench toolkit gives higher scores to slightly oversaturated colors, and this does not always align with the human preference.
\subsection{Details on deployments and on-device measurements} \label{sec:ondevice_supp}

Reported latencies in Table \ref{tab:ondevice_sup} are measured on research prototype devices with a
Qualcomm Snapdragon® 8 Gen. 5 Mobile Platform and
a HexagonTM processor, for running inference on the platform's NPU component.

As expected, DiT is the clear run-time bottleneck of the on-device pipeline. The High number of parameters, context lengths and consequently large activation maps lead to a strenuous edge inference use-case based on limited resources of the mobile environment. Thus, porting attempts of the entire base graph result in OOM, hence a split-based porting alternative was adopted, with relevant number of splits - blocks and run-times reported. We differentiate between "pure" run-time and initialization time of the graph, as the downside of the split-based approach is that it requires unloading - loading splits for subsequent passes. As an example, porting the $6\times5-$block Wan2.2 5B - Finetuned, requires $6\times5$-block$_\text{run\_time}$ + $5\times5$-block$_\text{init\_time}$; we do consider the initial split always loaded in memory.

Head-pruning brings about the first relaxation of the on-device strain, enabling the porting of $3\times10$-block splits. Comparing the $6\times5$ and $3\times10$ variants, we see that despite run-times being comparable, it is more beneficial to run on fewer splits from an initialization times perspective.

Finally, full recurrence unlocks the final - claimed runtime, further boosting mobile-friendliness by effectively suppressing context lengths with its recurrent formulation. Similar conclusions can be deduced by comparing full recurrence variants, only this time we are able to port the full 30-block graph, hence leading to no additional initialization overheads. Comparing similar split-level configurations, across Base - Pruned and full recurrence variants, we see that full recurrence performs best on both initialization and run-times. From conducted run-trace analysis, its superiority can be attributed to its mobile-friendliness. Based and pruned (to a smaller extent) graphs impose significantly heavier DMA traffic, for moving data in-out the NPU's tightly coupled memory, i.e. the Spill-Fill operations, rendering the graph entirely idle in certain periods waiting on the completion of these transfers. On the other hand, full recurrence manages to alleviate those transfers and leads to a better utilized execution graph. This is also reflected by the relevant Spill-Fill numbers.

\begin{table}[]
\centering
\renewcommand{\arraystretch}{1.05} 
\footnotesize
\begin{tabular}{llllllll}
\toprule
Component  & \#Splits & \#Blocks &   Init (ms)  & Run (ms)  & Total (ms) & RAM (GB) & Spill/Fill (GB) \\
\midrule
Text Encoder & - & - & - & 147.1 & 147.1 & 9.7 & 0.7 \\
Wan Decoder & - & - & - & - & - & OOM & - \\
LiteX2V Decoder & - & - & - & 72.6 & 72.6 & 1.4 & 1.8 \\
Our Decoder & - & - & - & 501. & 501. & 1.9 & 10.6 \\
\midrule
DiT - variants & & & & & & \\
\midrule
Wan2.2 5B - Finetuned & \multirow{3}{*}{6} & \multirow{3}{*}{5} & 11920.5 & 9127.8 & 21048.3 & 4.1 & 313.2 \\
\ \ + Head Pruning (23\%) &  &  & 10000. & 7413. & 17413 & 3.5 & 259.8 \\
\ \ \ \ + Full recurrence  &  &  & 1590.3 & 6498. & 8088.3 & 1.5 & 91.2 \\
\midrule
Wan2.2 5B - Finetuned & \multirow{3}{*}{3} & \multirow{3}{*}{10} & - & - & - & OOM & - \\
\ \ + Head Pruning (23\%) &  &  & 7520. & 7650 & 15170. & 5.2 & 294.\\
\ \ \ \ + Full recurrence  &  &  & 1083.8 & 6407.7 & 7491.5 & 2.7 & 89.4 \\
\midrule
Wan2.2 5B - Finetuned & \multirow{3}{*}{1} & \multirow{3}{*}{30} & - & - & - & OOM & - \\
\ \ + Head Pruning (23\%) &  &  & - & - & - & OOM & - \\
\ \ \ \ + Full recurrence  &  &  & 0. & 6650. & 6650. & 9.9 & 87.6 \\
\bottomrule \\
\end{tabular}
\caption{On device measurements, corresponding to processing of a single $81\times480\times832$ sample, on Snapdragon® 8 Gen. 5 NPU. $\#$Splits and corresponding $\#$Blocks are reported for split-based profilings, Peak RAM, along with initialization and run-times yielding the total run-time. Finally, Spill and Fill corresponds to data transfers from and back to NPU's tightly coupled memory respectively, to make room for graph operations.}
\label{tab:ondevice_sup}
\end{table}

\section{Broader impacts}
This paper presents a method for converting a server-scale text-to-video generative model into a system that runs offline on a mobile device. Such capabilities may increase access to generative video technologies and support useful applications in mobile creativity, education, and assistive tools.

However, broader accessibility also introduces potential risks. In particular, efficient on-device generation may reduce the cost of producing misleading or malicious synthetic media, including deepfakes, impersonation, and disinformation.

We believe these risks should be addressed through continued research on  robust digital watermarking and content authentication.

\section{The use of assets}
Datasets:\begin{itemize}
    \item In-house synthetic data generated by Wan2.1 14B -- no license.
    \item DAVIS -- GPL-3.0 license.
\end{itemize}
Models:\begin{itemize}
    \item Wan2.1 14B -- Apache License 2.0.
    \item Wan2.2-TI2V-5B -- Apache License 2.0.
    \item LightX2V lighttaew2\_2 -- Apache License 2.0.
    \item VBench toolkit -- Apache License 2.0.
\end{itemize}


\end{document}